%% file: main.tex
\documentclass{article}

\usepackage[preprint]{neurips_2025}


\input{math_commands/header}

\input{math_commands/math_commands}

\usepackage{fancyvrb}

\usepackage{ragged2e}  
\newcolumntype{L}{>{\RaggedRight\arraybackslash}X} 

\usepackage{tabulary}
 \usepackage{colortbl} 
 
\usepackage{tabularx}

\usepackage{amsmath,amssymb}         
\usepackage[most]{tcolorbox}   
\usepackage[utf8]{inputenc}
\usepackage{wrapfig}      

\usepackage{enumitem}

\usepackage{pifont}
\usepackage{fontawesome5}  

\hypersetup{
    colorlinks=true,
    citecolor=blue,
    linkcolor=blue,
}

\tcbuselibrary{skins}
\usepackage{lipsum} 
\tcbuselibrary{listings}

\newtcolorbox{promptbox}[1][]{%
  enhanced,
  colback=gray!5, 
  colframe=black!70, 
  coltitle=white,
  fonttitle=\normalsize\bfseries, 
  title={#1},
  bottomrule=0pt,
  toprule=0.5pt,
  leftrule=0pt,
  rightrule=0pt,
  arc=6pt,         
  outer arc=6pt,   
  left=10pt,
  right=10pt,
  top=6pt,
  bottom=6pt,
  boxsep=3pt,
  before skip=6pt,
  after skip=6pt,
  toptitle=2pt,
  bottomtitle=2pt,
  boxrule=0.5pt,
  width=\textwidth,
  nobeforeafter
}

\title{Bridging Formal Language with Chain-of-Thought Reasoning to Geometry Problem Solving}

\usepackage{amsmath}
\usepackage{graphicx}
\usepackage{tcolorbox}
\usepackage{xcolor}
\usepackage{hyperref}
\usepackage{tabularx}
\tcbuselibrary{skins}
\newcolumntype{L}{>{\raggedright\arraybackslash}X}
\usepackage{authblk}

\definecolor{DarkGreen}{RGB}{0,100,0}
\definecolor{DarkBlue}{RGB}{0,0,139}

\newcommand\nnfootnote[1]{%
  \begin{NoHyper}
  \renewcommand\thefootnote{}\footnote{#1}%
  \addtocounter{footnote}{-1}%
  \end{NoHyper}
}

\author[1]{\bf Tianyun Yang{$^*$}}
\author[2]{\bf Yunwen Li{$^*$}}
\author[2,1]{\bf Ziniu Li{$^*$}}
\author[2]{\bf Zhihang Lin}
\author[2,1,3]{\bf Ruoyu Sun}
\author[1,3]{\bf Tian Ding{$^\ddag$}}
\affil[1]{Shenzhen Research Institute of Big Data}
\affil[2]{The Chinese University of Hong Kong, Shenzhen}
\affil[3]{Shenzhen International Center for Industrial and Applied Mathematics}

\begin{document}

\maketitle

\vspace{-0.4cm}

\begin{abstract}
Large vision language models exhibit notable limitations on Geometry Problem Solving (GPS) because of their unreliable diagram interpretation and pure natural-language reasoning. A recent line of work mitigates this by using symbolic solvers: the model directly generates a formal program that a geometry solver can execute. However, this direct program generation lacks intermediate reasoning, making the decision process opaque and prone to errors. In this work, we explore a new approach that integrates Chain-of-Thought (CoT) with formal language. The model interleaves natural language reasoning with incremental emission of solver-executable code, producing a hybrid reasoning trace in which critical derivations are expressed in formal language. To teach this behavior at scale, we combine (1) supervised fine-tuning on an 11K newly developed synthetic dataset with interleaved natural language reasoning and automatic formalization, and (2) solver-in-the-loop reinforcement learning that jointly optimizes both the CoT narrative and the resulting program through outcome-based rewards. Built on Qwen2.5-VL-7B, our new model, named GF-Reasoner, achieves up to 15\% accuracy improvements on standard GPS benchmarks, surpassing both 7B-scale peers and the much larger model Qwen 2.5-VL-72B. By exploiting high-order geometric knowledge and offloading symbolic computation to the solver, the generated reasoning traces are noticeably shorter and cleaner. Furthermore, we present a comprehensive analysis of method design choices (e.g., reasoning paradigms, data synthesis, training epochs, etc.), providing actionable insights for future research. Our data and code are available at 
\url{https://github.com/TianyunYoung/GF-Reasoner}.

\end{abstract}

\nnfootnote{$^*$: Equal contribution.}
\nnfootnote{$\ddag$: Corresponding author. Email: \texttt{dingtian@sribd.cn}}

\vspace{-0.6cm}
\section{Introduction}
\vspace{-0.2cm}

Large Vision Language Models (LVLMs) have emerged as powerful tools for a wide range of applications, demonstrating impressive capabilities in tasks such as visual question answering, video understanding, etc.~\citep {li2023blip, liu2023visual, team2024gemini, hurst2024gpt, lu2024deepseek, bai2025qwen2}. Despite these rapid advances, LVLMs still exhibit significant weaknesses in reasoning ability~\citep{chen2021geoqa,lu2024mathvista, zhang2024mathverse}. When confronted with tasks requiring spatial or geometric reasoning, current models frequently produce inconsistent or incorrect results, substantially limiting their utility in 
practical applications~\citep{liu2023visual, gupta2023visual, yang2024thinking}. 

\begin{figure}[t]
    \centering
    \includegraphics[width=0.9\linewidth]{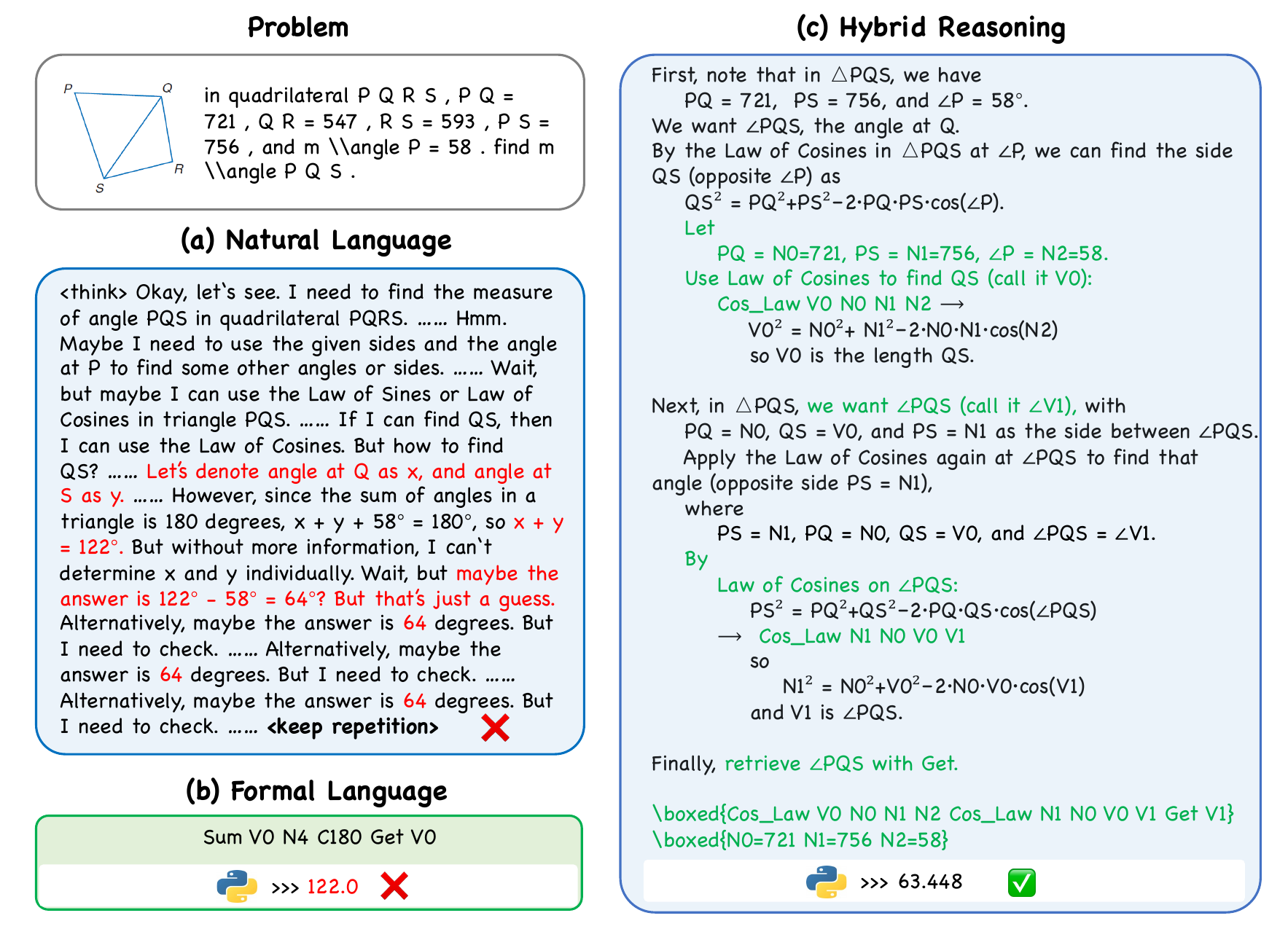}
    \caption{Illustration of formal-integrated reasoning for geometry problem solving. (a) Natural language solution from Vision-R1, in which the symbol ``..." denotes the omitted long reasoning steps; (b) Formal language solution generated by GeoX~\citep{xia2024geox}; (c) Hybrid reasoning solution produced by our GF-Reasoner model.}
    \label{fig:intro}
    \vspace{-0.2cm}
\end{figure}

This paper investigates Geometry Problem Solving (GPS) \citep{chen2023geoqa, lu2024mathvista}, a particularly challenging reasoning task where LVLMs must reason across geometric diagrams while applying geometry knowledge. Current research to improve LVLM reasoning capabilities has focused predominantly on \emph{natural language} reasoning approaches \citep{bai2025qwen2, team2025kimi, huang2025vision}, which, however, are limited in handling numerical computations, often produce redundant outputs, and cannot ensure the solution process is correct (see \cref{fig:intro}(a)). 

To tackle this, recent works have investigated the application of \emph{formal language}~\citep{lu2021inter,zhang2023formalgeo,zhang2023multi, xia2024geox}. Given textual questions and accompanying diagrams, LVLMs are trained to predict formal programs encoding solver-executable solution steps. The programs are subsequently executed by a geometry solver to derive numerical solutions (see \cref{fig:intro}(b)). However, the direct program generation approach lacks the flexibility in performing intermediate steps, which fundamentally restricts both the interpretability of its decision-making process and the model's reasoning capability. This raises a key research question: 
\vspace{-0.1cm}
\begin{center} 
\emph{How can we bridge intermediate reasoning with formal language in geometry problem solving?}
\end{center}
\vspace{-0.1cm}
In this work, we propose a hybrid reasoning framework that synergizes the flexibility of natural language reasoning with the precision of formal language. As shown in~\cref{fig:intro}(c), our framework interleaves natural language reasoning with the progressive generation of formal programs. Within this framework, natural language handles diagram interpretation, problem formalization, and reasoning trajectory planning. Meanwhile, formal programs implement geometric theorems as executable operators with explicit variable bindings during critical steps. This integration equips LVLMs with both general and rigorous reasoning capabilities for geometry problems.

How can we implement the above framework in practice? We find that simple prompting fails to achieve the desired results (see~\cref{sec:formal_language}), primarily because the formal language syntax and interleaved natural-formal reasoning pattern do not exist within the model's internal knowledge base. Therefore, we turn to post-training strategies—including Supervised Fine-Tuning (SFT) \citep{liu2024llava} and Reinforcement Learning (RL) \citep{guo2025deepseek,huang2025vision}—to teach the model to perform hybrid reasoning. This approach presents several technical challenges. First, such training data is scarce online, raising critical questions: How can we construct an effective hybrid reasoning dataset, and what essential characteristics should it possess? Second, determining proper training strategies remains challenging. Our experience indicates that some straightforward approaches fail to unlock hybrid reasoning potential and cannot match natural language reasoning performance.

To answer the above questions, we employ two scalable post-training strategies: 1) \emph{SFT on a newly developed Formal-Integrated Chain-of-Thought (FI-CoT) dataset.} We curate an 11k-sample CoT dataset featuring interleaved natural language and formal geometric language reasoning trajectories. The dataset is constructed via bidirectional synthesis to enhance the utilization efficiency of available data. SFT on this dataset enables the model to automatically formalize informal inputs into formats suitable for formal reasoning (e.g., by explicitly binding problem and process variables to operands) and internalize geometric formal language syntax. 2) \emph{Solver-integrated RL.} SFT alone proves insufficient for achieving strong formal-integrated reasoning capacity (see~\cref{sec:training}). Therefore, we develop a solver-in-the-loop RL framework where a geometric solver executes the resulting formal program and returns verification feedback on solution correctness. This iterative process refines the CoT trajectory and ultimately increases performance by a substantial margin.\footnote{Our empirical results show that RL increases Pass@1 accuracy by 26\%. Notably, without RL, our model fails to match the performance of natural language reasoning alone.}

\begin{figure}[t]
    \centering
    \includegraphics[width=\linewidth]{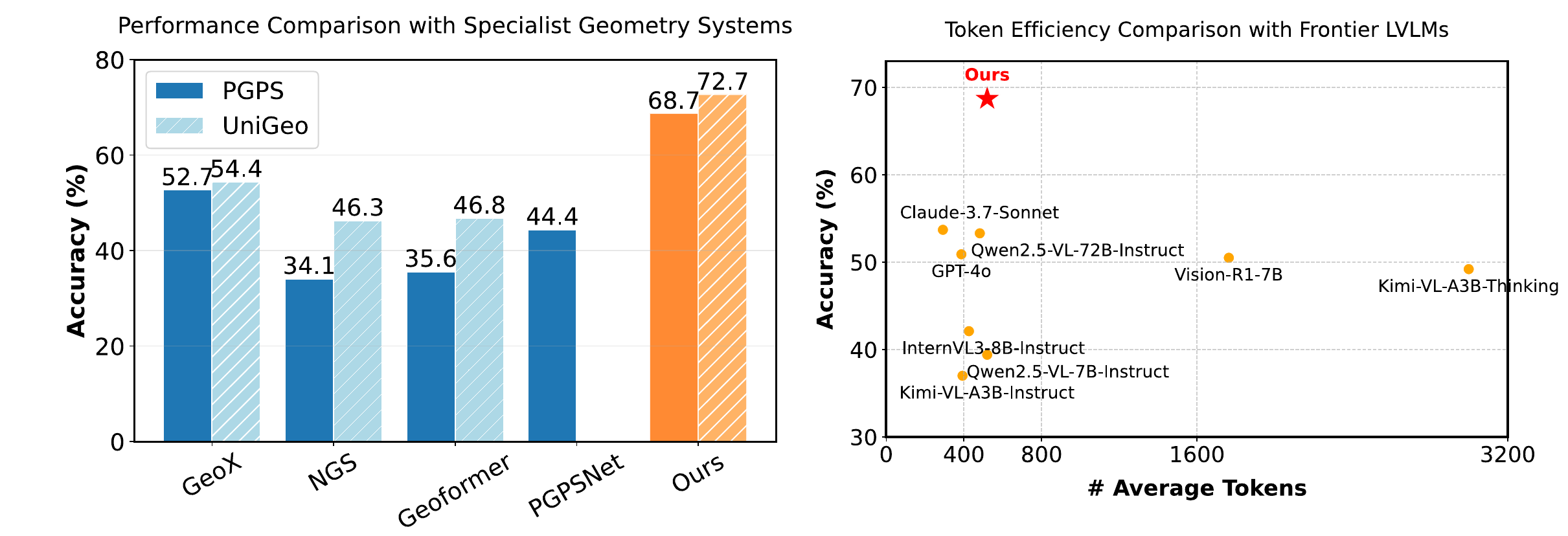}
    \caption{Left: Performance comparison with specialist geometry systems (GeoX, NGS, Geoformer, and PGPSNet) using direct formal program generation, evaluated on two benchmarks (PGSP and UniGeo). Right: Token efficiency comparison with frontier LVLMs.}
    \label{fig:model_comparison1}
    \vspace{-0.2cm}
\end{figure}

Built upon Qwen2.5-VL-7B-Instruct~\citep{bai2025qwen2}, our model, GF-Reasoner (Geometric Formal Reasoner), demonstrates superior performance over both specialized geometry systems (e.g., GeoX~\citep{xia2024geox}) and state-of-the-art multimodal LLMs (e.g., Qwen2.5-VL-72B~\citep{bai2025qwen2}), as shown in~\cref{fig:model_comparison1}.\footnote{PGPSNet is not evaluated on UniGEO as the benchmark lacks annotations for structural and semantic clauses required by the model.} We can also notice that our model has the additional benefits of concise outputs, achieved by condensing complex solving processes into compact operators and offloading explicit computation to an external solver. In line with these results,  we also present a comprehensive analysis of our framework designs spanning from the reasoning paradigm, data synthesis strategy, to training methodologies.

Our main contributions can be summarized as follows:
\begin{itemize}[topsep=1pt,parsep=1pt,partopsep=1pt, leftmargin=*]
    \item We propose a framework that integrates CoT with formal language and geometry solvers for GPS, a new hybrid reasoning paradigm that combines the strengths of both approaches.
    \item We curate a new 11k-sample formal-integrated CoT dataset using a bidirectional synthesis methodology. This is the first dataset of its kind, which enables us to explore post-training procedures such as SFT and RL to teach LVLMs external knowledge of geometric formal language.
    \item We empirically validate the effectiveness of integrating CoT with formal language, demonstrating improvements in both performance and token efficiency. We also provide comprehensive studies that may provide several actionable insights for future research.
\end{itemize}

\section{Related Work}

\textbf{Geometry Problem Solving (GPS).} GPS plays a crucial role in industry, manufacturing, and scientific areas. Since the 1970s, automated GPS has been an important research focus, which includes geometry calculation and geometry proving. In the context of geometry calculation, existing approaches fall into two categories: symbolic solvers and neural solvers. Symbolic solvers, such as GEOS~\citep{seo2015solving}, Inter-GPS~\citep{lu2021inter}, and E-GPS~\citep{wu2024gps}, parse diagrams and textual problems into several conditions. They then iteratively deduce new conditions by applying theorem rules until the problem is solved. While these solvers offer high interpretability, they rely on meticulously designed rules, making them difficult to extend. Recent advances in AI have shifted the focus from symbolic to neural methods. By leveraging neural networks to interpret geometry diagrams and derive formal program solutions, models like NGS~\citep{chen2021geoqa}, Geoformer~\citep{chen2022unigeo}, PGPSNet~\citep{zhang2023multi}, and GeoX~\citep{xia2024geox} have demonstrated promising results. These approaches use neural networks to parse the problem and predict a formal-language solution, which is then executed by an external solver to compute the numerical answer. 

Current neural methods have several notable limitations. On the one hand, they directly predict the result without intermediate reasoning, constraining their ability to solve more complex problems. On the other hand, some specialist geometry models face limited adaptability due to their restrictive input requirements. For example, GeoX,
NGS and Geoformer require problem variables to be explicitly declared in text questions (e.g., the text questions are like ``In triangle ABC, AC = N0, AB = N1, ...", where the value of N0 and N1 are predefined in a numerical list). In contrast to existing methods, our work is the first to integrate geometric formal language with CoT reasoning, possibly unlocking the reasoning potential of neural approaches for geometry calculation problems. Meanwhile, our method can auto-formalize the informal inputs by automatically binding problem variables into operands, offering higher flexibility.

\textbf{Multimodal Large Language Model Reasoning.} Improving complex reasoning capability of large models has been regarded as a critical pathway toward artificial general intelligence~\citep{guo2025deepseek}. In the field of large multimodal models, \citet{zhang2024mathverse} are among the first to systematically evaluate the mathematical reasoning ability of LVLMs. Their results reveal that LVLMs often generate incorrect answers with an incorrect reasoning process. Furthermore, among incorrect reasoning, calculation errors could contribute to 19.95\%. GeoSense \citep{xu2025geosense} is a recently developed benchmark to systematically evaluate the geometric reasoning abilities of LVLMs through the lens of geometric principles. They found that the identification and application of geometric principles remain a bottleneck for leading LVLMs. Recent advances in multimodal reasoning have led to several notable approaches~\citep{huang2025vision, xu2025visual, luo2025ursa, shen2025vlm, wang2025vl}. A representative example is Vision-R1~\citep{huang2025vision}, which employs DeepSeek R1 to automatically generate visual reasoning data from textual descriptions.  
While these methods demonstrate promising results, they primarily rely on natural language reasoning. In our work, we overcome natural language's imprecision and redundancy by combining it with formal language, enabling both flexible and concise geometric reasoning.

\section{Method}

This section presents our method for GPS. We begin by providing background on geometric formal language in \cref{sec:formal_language}, followed by our data synthesis approaches in \cref{sec:data_syn}, and finally present the training procedures in \cref{subsec:training_algorithms}.

\subsection{Geometric Formal Language}
\label{sec:formal_language}

Our work builds on the geometry solver and formal language defined in \citep{zhang2023multi}. It has 34 \textit{operators} and 55 \textit{operands}.\footnote{Compared to geometry solvers used in prior work \citep{chen2021geoqa,chen2022unigeo}, this formal language includes 16 additional operators and provides broader coverage of geometric theorems.} Each operator encodes a specific geometric theorem or axiom, covering fundamental operations across triangles, quadrilaterals, polygons, circles, and other geometric shapes. These operators mainly work with three types of operands: \textit{problem variables} (\texttt{N}) representing known measurements from the problem statement, \textit{process variables} (\texttt{V}) representing intermediate results generated during computation, and \textit{constants} (\texttt{C}) encoding common numerical values.

A formal program specifies a deduction step using these operators and operands. For example, the formal program $\boxed{\texttt{Gougu N0 N1 V0}}$ applies the Pythagorean theorem to calculate the hypotenuse, where \texttt{N0} and \texttt{N1} are two known leg lengths and \texttt{V0} is the hypotenuse to be computed. To leverage the geometry solver, the model must first understand the diagram and text question, define variables as operands (formalization), deduce relationships using geometry operators (reasoning), and finally construct the formal program (coding). After generating the formal program, both the program and extracted problem variables are sent to the geometry solver to compute numerical results for unknown process variables. This executor performs symbolic algebraic calculations to solve for unknown variables within the program's formulas, implemented via Python's SymPy library. Please refer to \cref{fig:intro} for illustration. 

This framework offers two notable advantages. First, it decouples reasoning from numerical computation by offloading arithmetic calculations to an external solver. This eliminates calculation errors that commonly plague natural language reasoning when involving complex computations (such as equation-solving, square root calculations, and inverse trigonometric functions that are common in GPS). Second, it provides significant conciseness advantages over natural language. For instance, while natural language requires multi-step descriptions (e.g., calculate the octagon's area by dividing it into triangles, computing each triangle's area, then summing the results), formal language can express the equivalent operation as a single line: $\boxed{\texttt{RNgon\_B\_Area C8 N0 V0}}$ (where \texttt{N0} = side length, \texttt{V0} = resulting area). This conciseness helps mitigate reasoning errors that arise from the long contexts required in verbose natural language reasoning.

\begin{figure}[t]
    \centering
    \includegraphics[width=1.0\linewidth]{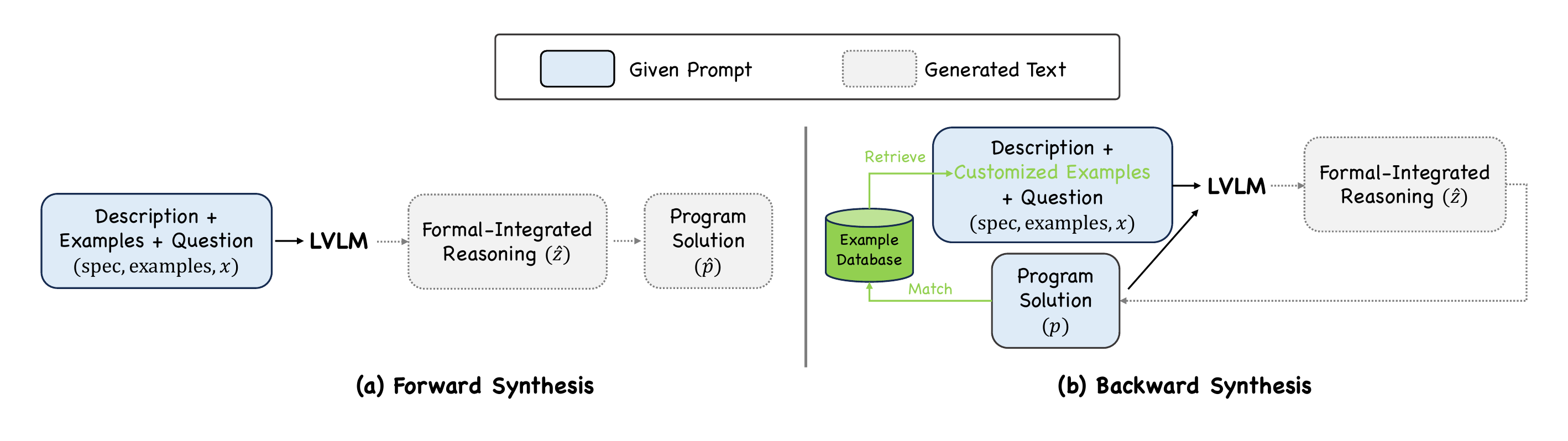}
    \caption{Overview of the formal-integrated CoT data synthesis process.}
    \label{fig:data_synthesis_overview}
\end{figure}

Our goal is to teach LVLMs to use this formal language and geometry solver. A straightforward approach involves designing prompt strategies that guide LVLMs in leveraging formal languages. However, our preliminary evaluation reveals significant limitations of this approach (see \cref{appendix:error_cases_prompting}): LVLMs, such as Qwen2.5-7B-VL-Instruct and its larger counterpart Qwen2.5-72B-VL-Instruct, frequently revert to natural language explanations instead of using geometry operators, or attempt to create operators not defined in the language specification. This indicates that existing LVLMs lack knowledge of geometric formal languages. This limitation may not be surprising, as geometry solver documentation and usage examples are unlikely to exist in the training corpora of existing LVLMs.

Given these limitations, we turn to effective and scalable post-training methods that update the model's parameters to incorporate the geometric formal language and solver. Two classes of learning approaches exist: 1) Supervised Fine-Tuning (SFT) using labeled formal language annotations~\citep{liu2024llava}, and 2) Reinforcement Learning (RL) where LVLMs generate formal language programs and receive feedback from solvers~\citep{guo2025deepseek, huang2025vision}. However, using RL alone is intractable because standard LVLMs lack knowledge of geometry solvers and cannot generate meaningful solver interactions for self-improvement. Therefore, we must first inject external tool usage knowledge into the model, typically through SFT. This creates a key technical challenge: the required SFT training data is not publicly available. We address this data scarcity problem in the following sections.

\begin{figure}
    \centering
    \includegraphics[width=1.0\linewidth]{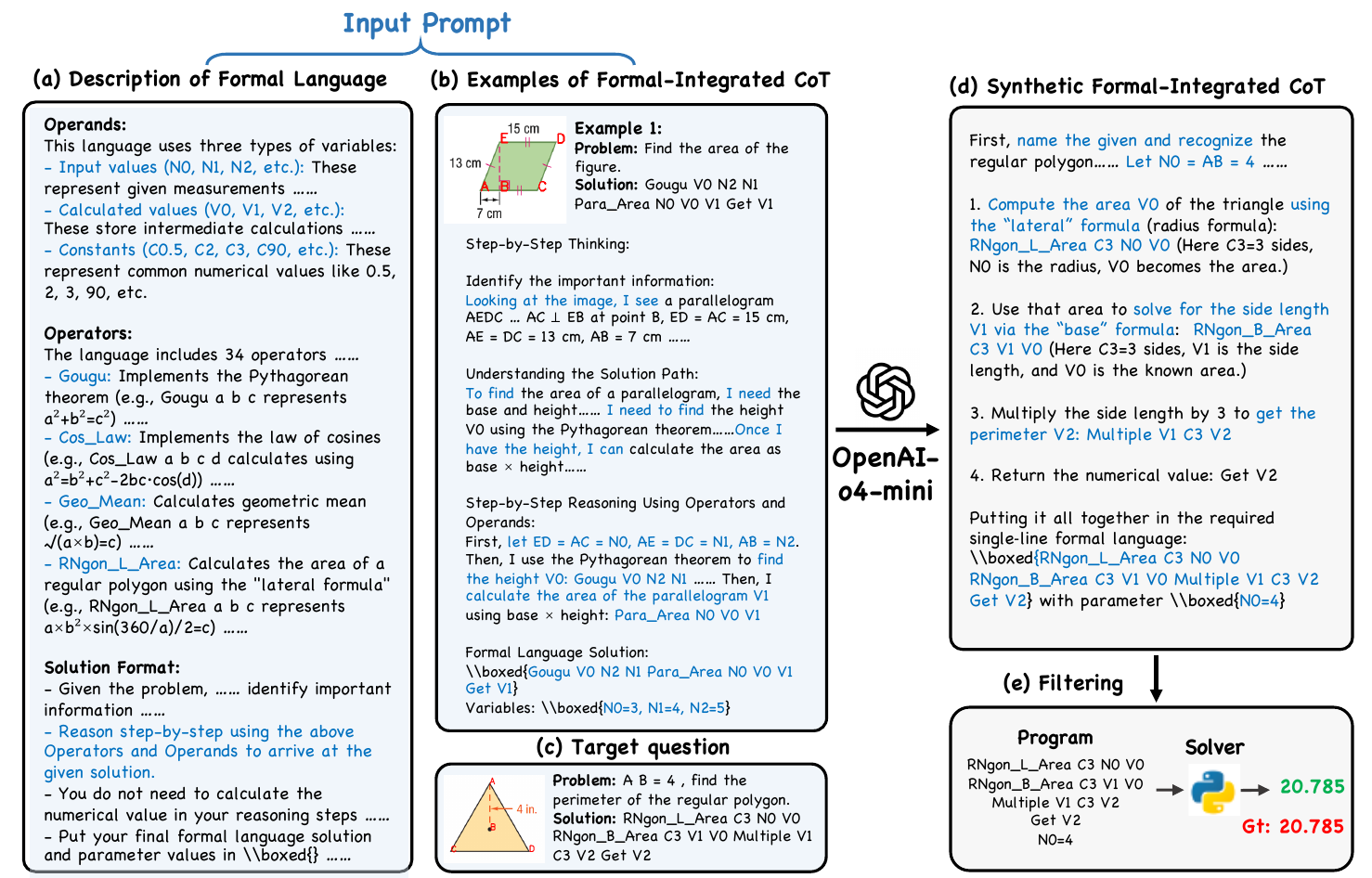}
    \caption{Detailed explanation of the backward synthesis process of formal-integrated CoT data.}
    \label{fig:data_synthesis}
\end{figure}

\subsection{Formal-Integrated CoT Data Synthesis}
\label{sec:data_syn}

In this section, we curate high-quality data to teach LVLMs how to perform reasoning with interleaved formal language. Our approach builds upon state-of-the-art LVLMs (e.g., Qwen-2.5-VL-72B and OpenAI-o4-mini) in our experiments—and leverages existing GPS datasets including PGPS9k~\citep{zhang2023multi}, UniGeo~\citep{chen2022unigeo}, and Geo170K~\citep{gao2023g}. While OpenAI-o4-mini lacks specialized knowledge of geometry solvers, it demonstrates exceptional capabilities in advanced reasoning, instruction following, and in-context understanding, making it well-suited for synthetic data generation within our framework.

Our core strategy involves providing comprehensive definitions of all operands and operators alongside detailed examples as demonstrations within the prompt to guide LVLMs. The operands and operators supply formal definitions, while the detailed examples illustrate how to combine reasoning with formal language to reach solutions. To illustrate, we formally define our synthesis setup using the following notations:
\begin{itemize}[topsep=1pt,parsep=1pt,partopsep=1pt, leftmargin=*]
    \item $\texttt{spec}$: Detailed description of all formal operands and operators in~\cref{sec:formal_language} and solution format requirements.
    \item $\texttt{examples} = { \langle x_i,z_i,p_i \rangle_{i=1}^{n} }$: Set of manually written example demonstrations showing formal-integrated CoT reasoning using geometric symbols.
    \item $x$: Target question to be solved, containing the geometric problem statement and diagram.
    \item $z$: Synthetic formal-integrated CoT reasoning trajectory generated by the state-of-the-art LVLM.
    \item $p
    $: Final formal program representing the executable solution. 
\end{itemize}

The examples demonstrate a response style that first formalizes the problem using both problem variables and process variables, then explicitly cites relevant operators and theorems to derive the final solution. This approach aligns with the deliberative alignment approach in \citep{guan2024deliberative}.

\textbf{Synthesis Objective.} Given the structured prompt containing formal description, demonstration, target question, and the ground-truth program (when available), our goal is to generate high-quality formal-integrated reasoning $z$ that: 1) demonstrates step-by-step problem solving using geometric operators from the description, (2) produces a valid formal program that can be executed by the geometric solver. To achieve this, we design two complementary synthesis strategies tailored to different dataset types, as illustrated in~\cref{fig:data_synthesis_overview}. For datasets containing only ground-truth numerical solutions without formal program annotations (e.g., UniGEO, Geo170k), we utilize \textit{forward synthesis} to generate both reasoning steps and formal program solutions from the given problems. For datasets with ground-truth formal program annotations (e.g., PGPS9k), we employ \textit{backward synthesis} to derive reasoning trajectories by backtracking from problems to their formal program solutions. The synthesis prompts are shown in~\cref{appendix:prompt}.

\textbf{Forward Synthesis.} Given the structured prompt $(\texttt{spec}, \texttt{examples}, x)$, the LVLM will generate $\widehat{y} = [\widehat{r}, \widehat{p}]$, where $\widehat{z}$ represents the model-generated formal-integrated CoT reasoning trajectory and $\widehat{p}$ denotes the corresponding formal program solution. To filter out valid synthetic samples with correct reasoning chain $\widehat{z}$ and program $\widehat{p}$, we execute $\widehat{p}$ with a geometric solver and compare its output to the ground-truth numerical solution.

A key advantage of this approach is its flexibility, requiring only the question prompt and ground-truth numerical solution - no ground-truth program $p$ is needed. However, the success rate of forward data synthesis is relatively low. Our preliminary experiments show that Qwen-2.5-VL-72B achieves less than 20\% success rate. These results highlight the difficulty of establishing formal language reasoning ability through prompting alone. To address this limitation, we observe that some existing datasets (e.g., PGPS9K) already contain labeled programs $p$, which can serve as valuable hints for LVLMs during data synthesis, leading to the backward synthesis approach described below.

\textbf{Backward Synthesis.} We develop another backward synthesis method using the extended prompt $(\texttt{spec}, \texttt{examples}, z, p)$ that includes the correct program solution. We task the model with generating the reasoning chain $\widehat{z}$ that leads to the given $p$. We design a customized backward synthesis strategy to retrieve corresponding in-context learning examples from a manually written example database by matching the first operator in the solution program. This method further improves synthesis accuracy (shown in \cref{tab:data_synthsis} later). The improvement stems from the model's prior knowledge of the correct program, enabling it to concentrate on reconstructing the reasoning path that links the question to the formal solution.

In total, we synthesize 11k formal-integrated CoT reasoning samples, comprising 2.2k and 3.8k samples generated through forward data synthesis on the UniGEO and Geo170K training sets, respectively, and 5k samples produced via backward synthesis on the PGPS9K training set. An example generated by backward synthesis is shown in~\cref{fig:data_synthesis}.


\subsection{Training Procedures}
\label{subsec:training_algorithms}

\begin{figure}
    \centering
    \includegraphics[width=0.95\linewidth]{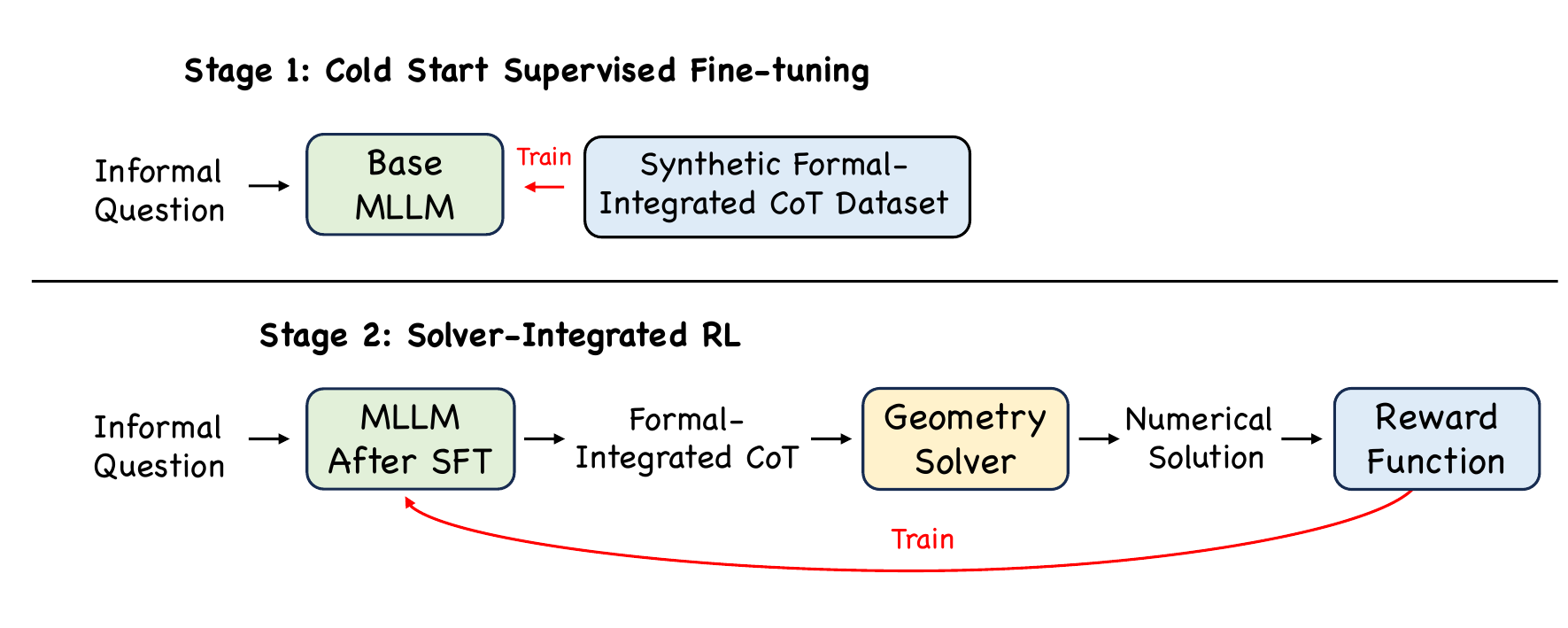}
    \caption{The two-stage training pipeline of our framework.}
    \label{fig:training}
\end{figure}

\textbf{Stage 1: Cold-Start Supervised Fine-tuning.} Using the synthetic reasoning data, we can conduct supervised fine-tuning of the LVLM (e.g., Qwen-VL-2.5-7B-Instruct) to establish foundational formal-integrated reasoning capabilities. The training objective is:
\begin{align}
\min_{\theta} \mathbb{E}_{(x,z,p) \sim \mathcal{D}}[-\log \pi_{\theta}(z,p|x)]
\end{align}
where $\mathcal{D}$ denotes our synthetic dataset containing geometry problems paired with formal-integrated reasoning chains and programs, and $\pi_\theta$ is the probability distribution of the LVLM. This initial stage serves as a warm-up phase, equipping the model with two fundamental capabilities: (1) autoformalization of informal inputs into a format suitable for formal reasoning, and (2) basic comprehension of geometric formal language syntax. We empirically find that SFT alone is insufficient to achieve strong performance given the training samples and compute budget available to us. One possible explanation is that training with offline data constrains the model's behaviors. To address this limitation, we explore RL with online self-generated data to improve generalization performance.

\textbf{Stage 2: Solver-Integrated Reinforcement Learning.} After training on reasoning samples in the first stage, the model possesses effective exploration capability and basic formal language reasoning ability, though these abilities are not yet perfect. The subsequent reinforcement learning stage further refines these abilities through solver-guided optimization. In this phase, the model interacts with a geometric solver that provides verification results for the output program, enabling trial-and-error learning. The objective is as follows:
\begin{align}
\max_{\theta} \mathbb{E}_{(z, p) \sim \pi_{\theta}(\cdot|x)}[r(z,p)],
\end{align}
where the reward function $r(z, p)$ is computed by executing the generated program $p$ through the geometric solver to obtain a numerical result for the target variables specified in the question, where $r(r,p)=1$ if the result matches the ground-truth solution, and $0$ otherwise. Following \citep{li2024remax}, we opt for the value-model-free RL algorithms with direct policy gradients for the above reward maximization, due to its efficiency and theoretical suitability. Specifically, we choose the GRPO algorithm~\citep{shao2024deepseekmath} with a higher clip ratio~\citep{yu2025dapo}.

\section{Experiments}


In this section, we present experiments that justify our framework. We will first present the main results, comparing our best-configured model with frontier models on GPS benchmarks. Then, we provide the  error analysis and case studies, providing insights about what behaviors are unique when leveraging formal language and geometry solvers, and their advantages over natural language reasoning. Finally, we present comprehensive controlled experiments showing that achieving the above performance is not trivial, since some straightforward approaches fail to beat or even match SOTA performance. These experiments may provide actionable insights into critical designs and practices that  guide future work.

\subsection{Experiment Setup}

\textbf{Implementation Details.} We initialize training using the synthetic formal-integrated CoT data on Qwen-2.5-VL-7B-Instruct, implemented via the LLaMA-Factory framework, running for 2 epochs on 4 NVIDIA A100 80GB GPUs. Following this cold-start phase, the model undergoes further RL optimization via the GRPO algorithm~\citep{shao2024deepseekmath} implemented in the Verl framework. The model is trained on training subsets from PGPS9k, UniGEO, and Geo170k using a geometry solver-based reward function that compares the numerical result obtained from executing the generated programs against ground truth solutions. We run for 15 epochs on 8 NVIDIA H20 96GB GPUs.

\textbf{Baselines.} We compare our method against three categories of baselines: 1) Specialist geometry systems for geometry calculation problems, comprising GeoX~\citep{xia2024geox}, NGS~\citep{chen2021geoqa}, Geoformer~\citep{chen2022unigeo}, and PGPSNet~\citep{zhang2023multi}. 2) Open-Source General LVLMs, which include open source models Qwen2.5-VL-7B~\citep{bai2025qwen2}, Qwen2.5-VL-72B, InternVL3-8B-Instruct~\citep{zhu2025internvl3}, Vision-R1-7B~\citep{huang2025vision}, Kimi-VL-A3B-Instruct~\citep{team2025kimi}, and Kimi-VL-A3B-Thinking. 3) State-of-the-art Closed-Source LVLMs including GPT-4o (2024-11-20) and Claude-3.7 Sonnet (2025-02-19).

\textbf{Datasets.} We assess geometric reasoning capabilities across four established benchmarks: (1) the test splits of PGPS9K and UniGEO, and (2) the plane geometry subsets from MathVista~\citep{lu2024mathvista} and MathVerse~\citep{zhang2024mathverse} testmini splits, which provide cross-domain evaluation of mathematical visual reasoning.

\textbf{Evaluation Metrics.} Performance is measured by comparing numerical solutions against ground-truth answers, reporting the Pass@1 accuracy. To eliminate choice bias when evaluating on MathVista and MathVerse, we reformulate multiple-choice questions as free-form generation questions, ensuring fair assessment of the model's inherent reasoning ability.

\begin{table}[t]
\centering
\caption{Performance comparison with fountier LVLMs.}
\label{tab:model_comparison2}
\setlength{\tabcolsep}{0.4em} 
\begin{tabular}{lccccc}
\toprule
Model & Params & PGPS & UniGeo & MathVista & MathVerse \\
\midrule
GPT-4o (2024-11-20) & - & 50.9	&43.9	&47.1 & 43.3 \\
Claude-3.7 Sonnet (2025-02-19) & - & 53.7	& 47.2&  52.4& 47.2  \\
Qwen2.5-VL-7B-Instruct~\citep{bai2025qwen2} & 7B & 39.4 & 51.5 & 52.4 & 39.9 \\
Qwen2.5-VL-72B-Instruct~\citep{bai2025qwen2} & 72B & 53.3 &	67.9 &	63.0 &	52.1 \\
Vision-R1-7B~\citep{huang2025vision} & 7B & 50.5 & 60.9 &	59.1 &	39.9 \\
InternVL3-8B-Instruct~\citep{zhu2025internvl3} & 7B & 42.1	& 50.0	& 50.5	& 38.7 \\
Kimi-VL-A3B-Instruct~\citep{team2025kimi}  & 16B & 37.0	& 42.2 & 42.8 & 36.4 \\
Kimi-VL-A3B-Thinking~\citep{team2025kimi}  & 16B & 49.2	& 48.7 & 62.5 & 44.1 \\	
\midrule
\rowcolor{blue!20} \textbf{GF-Reasoner (Ours)} & 7B & \textbf{68.7} & \textbf{72.7} & \textbf{64.9} & \textbf{52.2} \\
\bottomrule
\end{tabular}
\end{table}

\subsection{Main Results}

\textbf{Comparison with Specialist Geometry Systems.} \cref{fig:model_comparison1} (left) demonstrates our method's performance on geometric problem solving compared to specialist geometry systems.\footnote{The baseline scores are sourced from \citep{xia2024geox}.} Our method achieves significantly better performance than existing geometry solvers, with gains of +16 points on PGPS9K and +18.3 points on UniGEO. Additionally, it is noteworthy that our method exhibits much higher flexibility by eliminating the need for pre-formalized questions and additional clauses. It can extract problem variables and formalize informal questions directly in the reasoning process. In contrast, specialist systems have limited adaptability due to their restrictive requirements: (1) GeoX, NGS, and Geoformer require problem variable formalization within the text question.  (2) PGPSNet requires additional structured and semantic clauses beyond the text question and image diagram, which are typically unavailable in standard geometry problems.

\textbf{Comparison with Frontier LVLMs.} \cref{tab:model_comparison2} demonstrates our method's performance compared with existing frontier LVLMs released this year. Among the evaluated baselines, we notice that Qwen2.5-VL-72B-Instruct achieves the strongest baseline results. Compared with it, our method achieves consistent superiority across evaluated benchmarks, achieving significant performance gains of +15.2 points on PGPS9K and +4.8 points on UniGEO. These improvements highlight our approach's superiority in multiple domains of geometry solving problems. 

The integration of formal language also enhances token efficiency, enabling more concise and precise reasoning processes. As demonstrated in~\cref{fig:model_comparison1} (right), among open-source models with fewer than 16B parameters, thinking models (e.g., Vision-R1-7B, Kimi-VL-A3B-Thinking) achieve higher accuracy at the cost of increased token consumption, while short-CoT models (e.g., Qwen2.5-VL-7B) prioritize token efficiency but sacrifice accuracy. Our formal-integrated model breaks this trade-off by low token usage while simultaneously improving geometry problem-solving accuracy by 15\% over the most effective baseline (Claude-3.7-Sonnet). This token efficiency gain stems from formal language's capacity to eliminate redundant natural language explanations.

\subsection{Analysis}
\label{sec:analysis}

In this section, we conduct ablation studies to systematically evaluate our framework's key components. Our analysis focuses on three critical aspects that contribute to the overall performance: (1) reasoning paradigm, particularly the benefits of bridging formal language with CoT reasoning in test-time scaling\footnote{Test-time scaling refers to leveraging additional computational resources during inference to enhance model performance \citep{snell2024scaling}.} and error reduction, (2) data synthesis strategies for training data preparation, (3) SFT and RL training recipes for establishing foundational knowledge and strategy refinement.

\subsubsection{Benefits of Formal Language Integration}
\label{sec:analyze_benefit}

\begin{figure}[t]
    \centering
    \begin{subfigure}[b]{0.32\linewidth}
        \includegraphics[width=\linewidth]{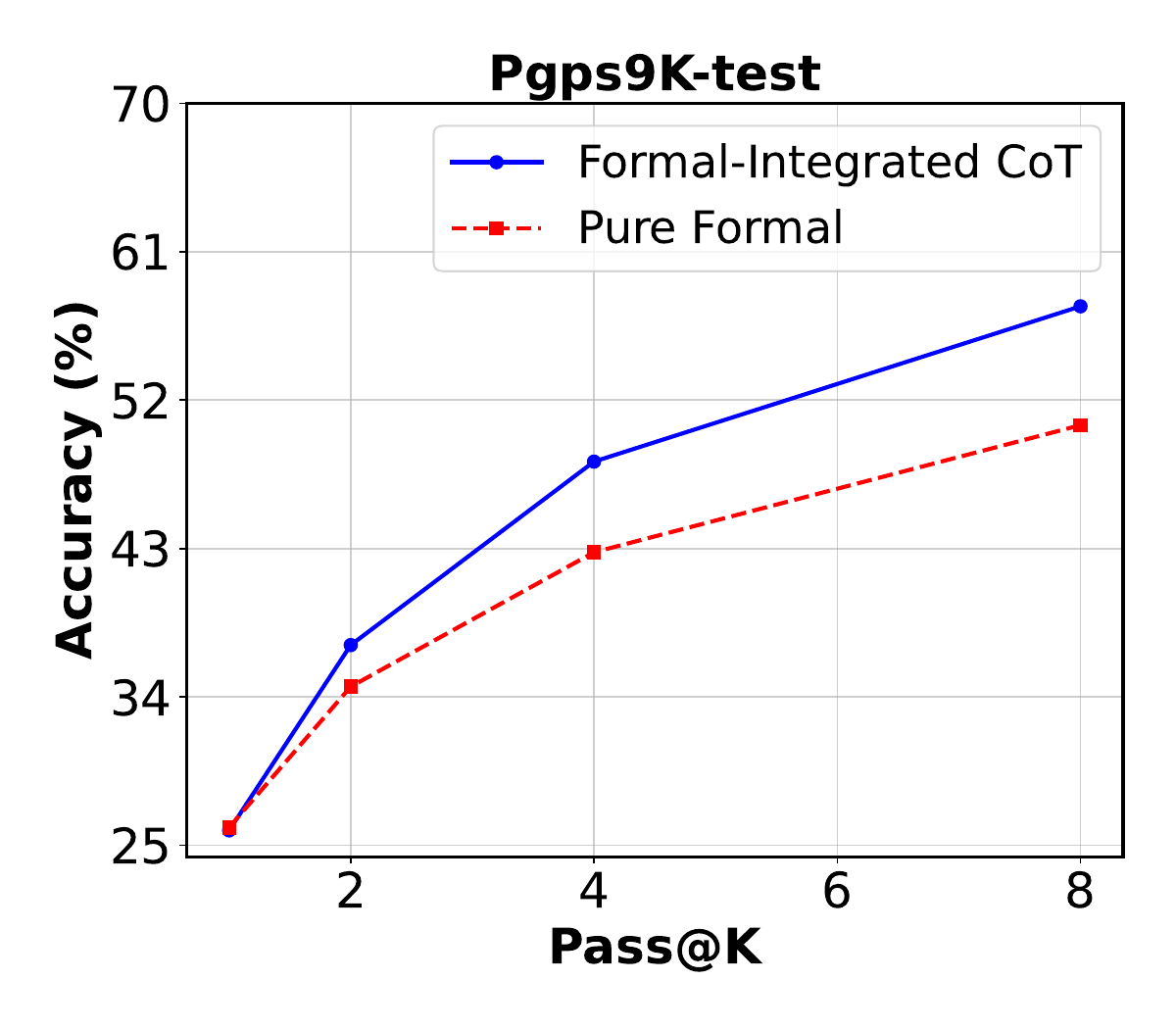}
    \end{subfigure}
    \hfill
    \begin{subfigure}[b]{0.32\linewidth}
        \includegraphics[width=\linewidth]{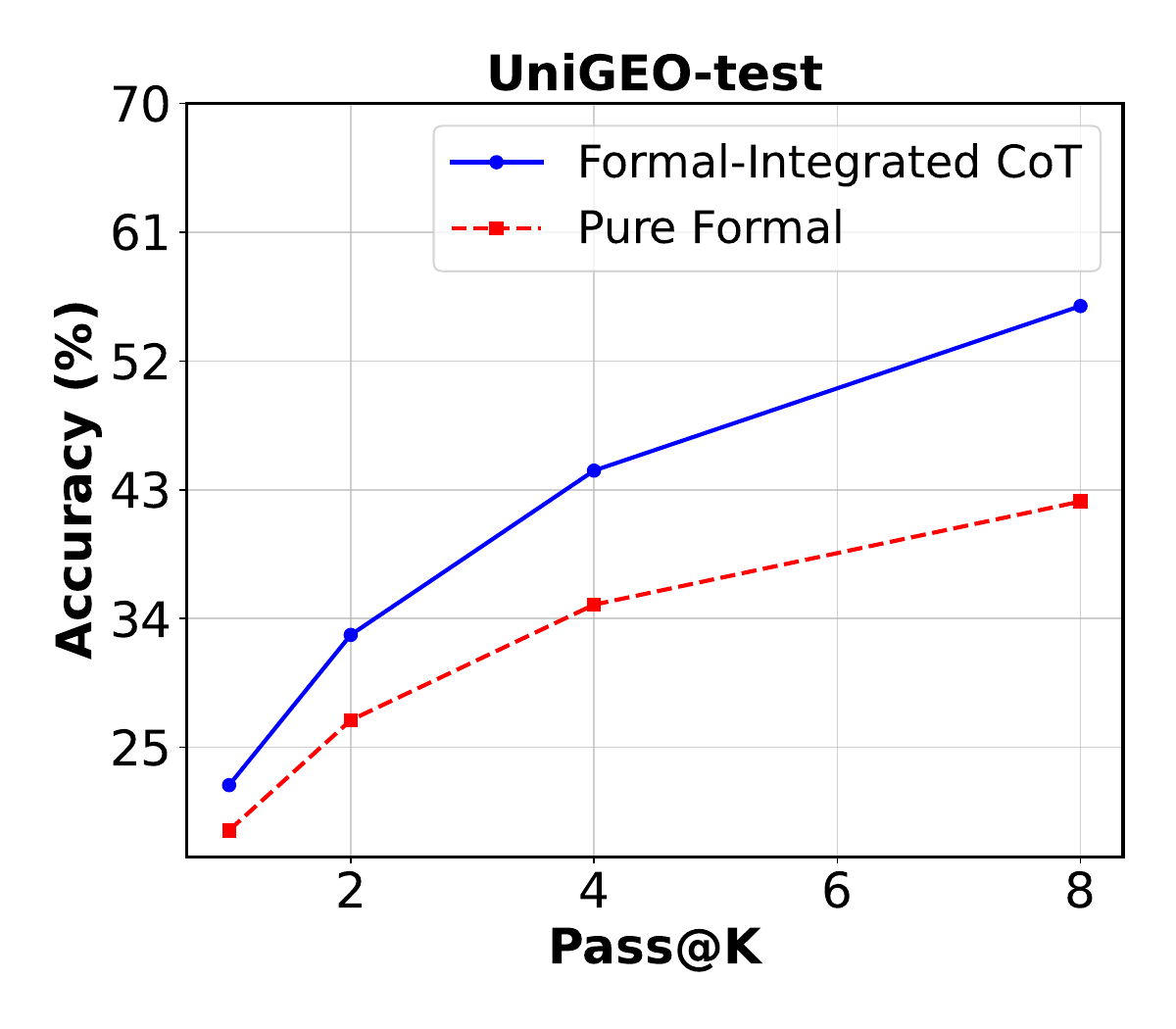}
    \end{subfigure}
    \hfill
    \begin{subfigure}[b]{0.32\linewidth}
        \includegraphics[width=\linewidth]{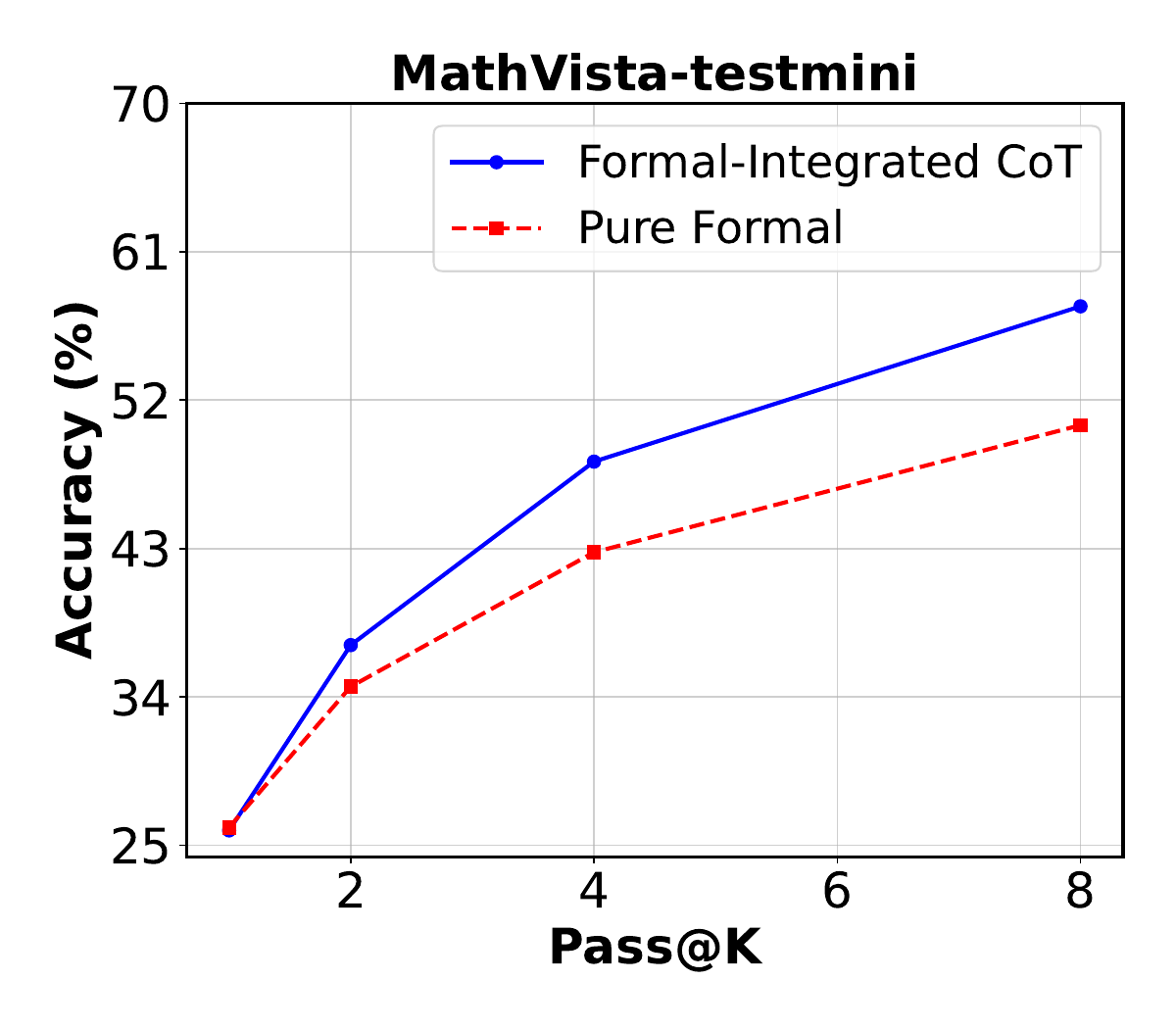}
    \end{subfigure}
    \caption{Pass@K performance of supervised fine-tuned model with and without CoT reasoning.}
    \label{fig:pass_k}
\end{figure}

\textbf{Enhanced Inference-Time Performance through Bridging Formal Language with CoT.} We conduct a controlled study comparing bridging formal language with CoT reasoning against direct formal language prediction. We perform SFT on the Qwen-VL-2.5-7B-Instruct model using two carefully curated datasets derived from PGPS9K, matched in size and sample composition. The first dataset includes responses with step-by-step formal-language-interleaved reasoning trajectories, while the second contains only final formal program outputs. As shown in~\cref{fig:pass_k}, integrating formal language with CoT reasoning demonstrates superior Pass@K scaling, with the performance gap widening as the number of samples increases. Notably, on the PGPS9k dataset, CoT and non-CoT achieve comparable performance in Pass@1 evaluation, while CoT achieves a 7\% relative improvement in Pass@8. This evidence suggests CoT's intermediate reasoning steps enable more effective exploration of the solution space during inference. Our finding aligns with \citep{prystawski2023think}, demonstrating that CoT enables the chaining of local knowledge to estimate relationships between variables not observed together during training.

\begin{table}[t]
\centering
\caption{Results on four types of errors.}
\label{tab:error_percent_compare}
\begin{tabular}{lcc}
\toprule
\setlength{\tabcolsep}{2em}
\textbf{Error Type} & \textbf{Natural Reasoning} & \textbf{Formal-Integrated Reasoning}\\
\midrule
Reasoning Error& 23.0\% & \textbf{14.3\%} \\
Geometry Knowledge Error & 12.3\% & \textbf{10.3\%} \\
Computation Error & 1.7\% & \textbf{0.3\%} \\
Visual Perception Error & \textbf{3.0\%} & 8.0\% \\
\midrule
All & 40\% & \textbf{32.9\%} \\
\bottomrule
\end{tabular}
\end{table}

\textbf{How Integrating Formal Language in Reasoning Reduces Errors?} To evaluate the advantages of integrating formal language in reasoning over traditional natural language reasoning, we categorize errors into four distinct types and evaluate how each type of error is reduced. The error type categorization follows existing works~\citep{zhang2024mathverse,lu2024mathvista}: 1) visual perception errors, 2) reasoning errors, 3) geometric knowledge errors, and 4) computation errors. Representative examples of each error type are shown in~\cref{appendix:error_catogory}.

To provide a comparative baseline for solving the geometry problems in natural language reasoning, we trained another model using the Qwen-VL-2.5-7B-Instruct base model with reinforcement learning on the same dataset as our model. The reward function was calculated by comparing the extracted numerical solution from the generated response with the ground truth solution. 

We conducted an error analysis on the same problem set (300 samples from PGPS9K-test) to highlight the differences between our formal-integrated model and the natural language baseline model. The error analysis, presented in~\cref{tab:error_percent_compare}, demonstrates reductions in reasoning errors, geometry knowledge errors, and computation errors. Notably, formal-integrated reasoning reduces reasoning errors by 8.7\%. Besides, by offloading symbolic computation to an external solver, formal-integrated reasoning reduces computation errors to nearly zero despite the already low baseline computation error rate.\footnote{The computation error in formal-integrated reasoning is not zero because there is one sample where the model incorrectly calculated intermediate numerical results in natural language without using formal language.} For concrete examples of how formal reasoning eliminates computation and reasoning errors, refer to~\cref{appendix:error_reduction_case}. We can observe that formal-integrated reasoning offers a more reliable and precise reasoning pathway than pure natural language. Besides,

\subsubsection{Data Synthesis Strategies}
\label{sec:analysis_data}

\textbf{Customized Backward Synthesis for Improved Synthesis Accuracy.} Table~\ref{tab:data_synthsis} compares three synthesis strategies (forward, backward, and customized backward) using Qwen-VL-2.5-72B. We evaluate synthesis accuracy by extracting both the formal solution program and problem variables, then verifying them against ground truth numerical solutions using a geometry solver.

\begin{wraptable}{r}{0.4\textwidth} 
\centering
\setlength{\tabcolsep}{0.2em} 
\caption{Comparison of different synthesis strategies.}
\begin{tabular}{lc}
\toprule
\textbf{Method} & \textbf{Accuracy(\%)} \\
\midrule
Forward & 19.5\\
Backward & 47.5 \\
Customized Backward & \textbf{50.0}\\
\bottomrule
\end{tabular}
\label{tab:data_synthsis}
\end{wraptable}

Our results demonstrate that backward synthesis yields a significant accuracy improvement (19.5\% → 47.5\%). This suggests that reconstructing the reasoning path from question to formal solution is more effective when the solution destination is known. Further performance gains (47.5\% → 50.0\%) are achieved by customizing in-context learning (ICL) examples based on the target solution program, highlighting the importance of example-question alignment for efficient synthesis.

\begin{figure}[t!]
    \centering
    \begin{subfigure}[b]{0.3275\linewidth}
        \includegraphics[width=\linewidth]{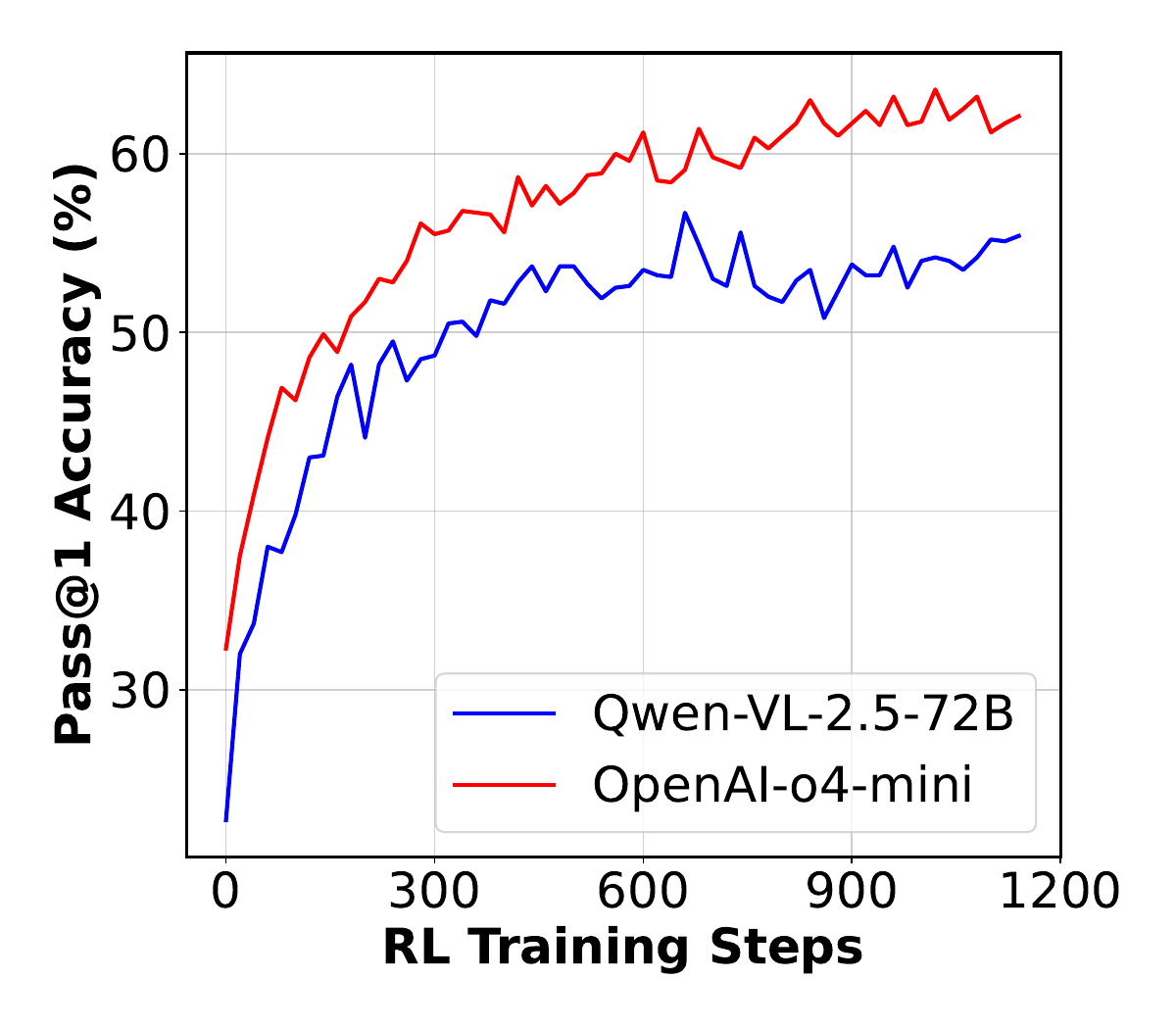}
        \caption{Style}
        \label{fig:style}
    \end{subfigure}
    \begin{subfigure}[b]{0.3275\linewidth}
    \includegraphics[width=\linewidth]{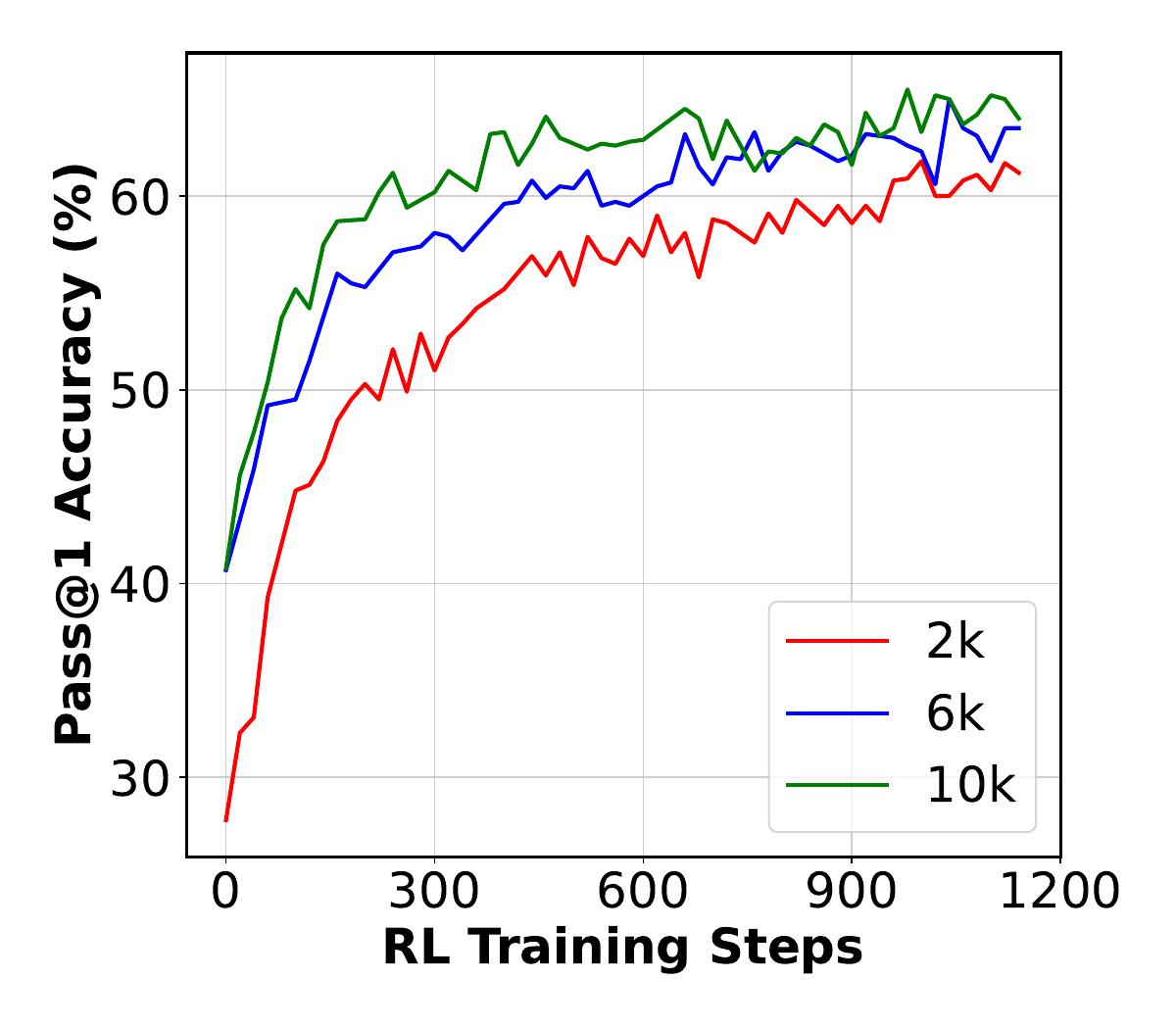}
    \caption{Amount}
        \label{fig:amount}
    \end{subfigure}
    \begin{subfigure}[b]{0.3275\linewidth}
        \includegraphics[width=\linewidth]{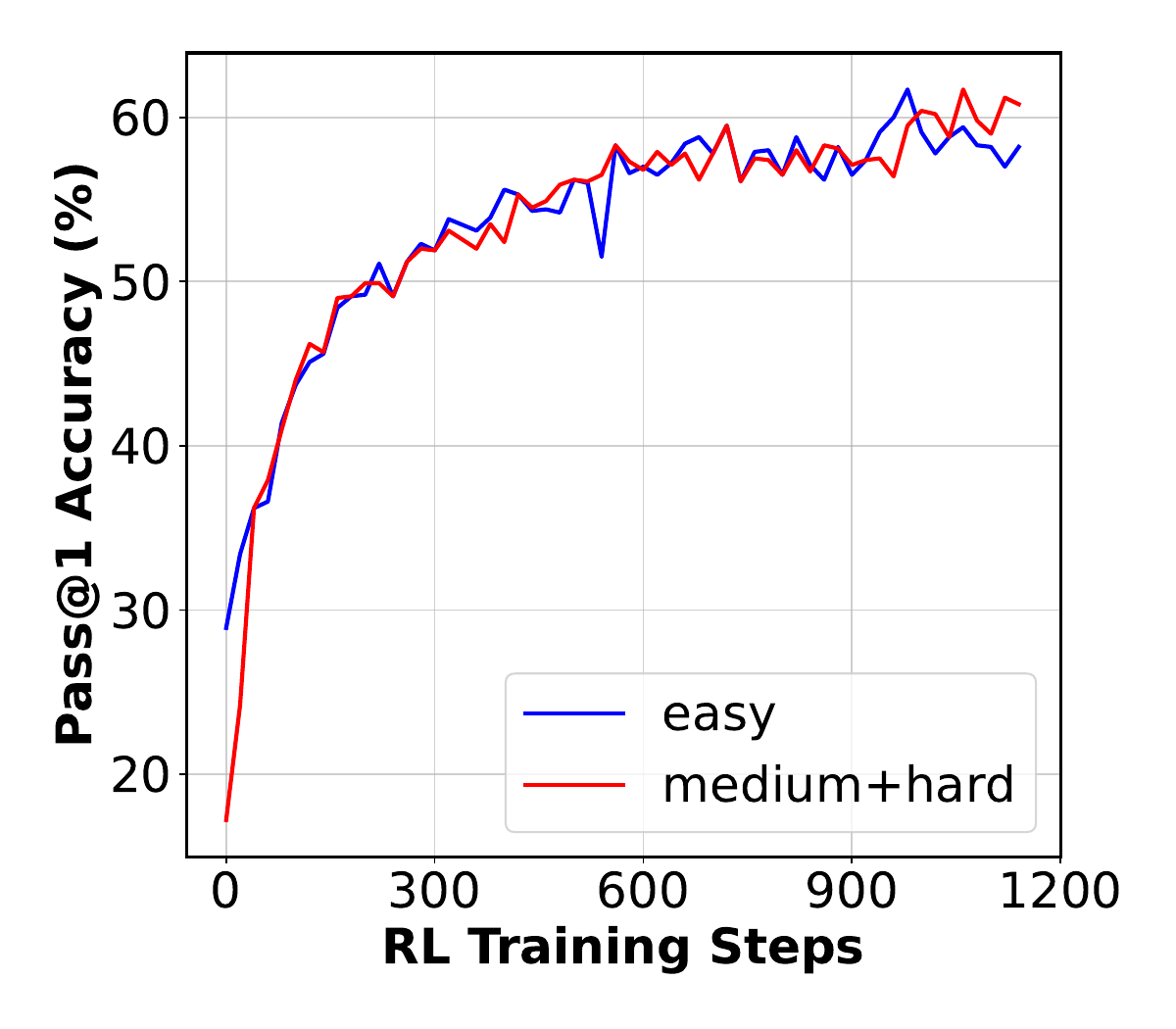}
        \caption{Difficulty}
        \label{fig:difficulty}
    \end{subfigure}
    \caption{Performance on the PGPS9K test set during RL training varying SFT initialization dataset, including CoT trajectory style, dataset amount, and difficulty coverage. }
    \label{fig:data_ablation}
\end{figure}

\textbf{Impact of SFT Training Data Characteristics.} The quality and composition of formal-integrated CoT data during SFT training directly influence the base model's mastery in the formal language for geometry problem solving, and consequently affect reasoning trajectory refinement in RL training. We systematically investigate three dimensions of formal-integrated data, analyzing several key variants:
\begin{itemize}[topsep=1pt,parsep=1pt,partopsep=1pt, leftmargin=*]
    \item \textbf{CoT Trajectory Style}: We compare the base models trained on formal-integrated reasoning trajectories generated by OpenAI-o4-mini and Qwen2.5-VL-72B, respectively, both are synthesized based on PGPS9K training set with an equal amount (3k each) and identical problem coverage. 
    \item \textbf{Dataset Amount}: With the OpenAI-o4-mini CoT style, we experiment by varying the number of training examples used, specifically using 2k, 6k, and 10k randomly sampled examples as SFT training data, respectively. 
    \item \textbf{Difficulty Coverage}: According to the number of geometry operators used, we categorize the training dataset into two difficulty levels, including easy ($\leq$ 3 operators) and medium/hard ($\geq$ 4 operators). We create equal-sized subsets for each difficulty level (2k each) and train the base model on the two subsets.
\end{itemize}

We perform the ablation study varying each of the three dimensions independently to isolate their impact on model performance. Results are presented in~\cref{fig:data_ablation}. 

The results in~\cref{fig:data_ablation}(a) demonstrate that RL training initialized with OpenAI-o4-mini-style CoT data consistently outperforms training based on Qwen-VL-2.5-72B-style data. To investigate this discrepancy, we analyze the reasoning trajectories of both models. We observe that OpenAI-o4-mini generates more logically consistent and formally precise reasoning steps, leading to better policy initialization for RL. Additionally, the results in~\cref{fig:data_ablation}(b) indicate that increasing the amount of SFT initialization data improves the final RL performance, but the gains diminish beyond a certain threshold (e.g., 6k samples). This suggests that a relatively small but well-curated dataset is sufficient for the model to learn the fundamental grammar and usage of geometric formal language during SFT. Additional data provides only marginal improvements. Interestingly, from~\cref{fig:data_ablation}(c), we could also observe that the difficulty level of SFT data (easy vs. medium/hard) does not significantly affect final RL performance. Although models trained on medium/hard CoT data initially underperform those trained on easy data, both converge to similar accuracy after RL training. This implies that SFT primarily serves to teach foundational knowledge of formal language, which can be acquired from either easy or hard problems, while RL compensates for higher-level reasoning gaps.

\begin{figure}
    \centering
    \begin{subfigure}[b]{0.32\linewidth}
        \includegraphics[width=\linewidth]{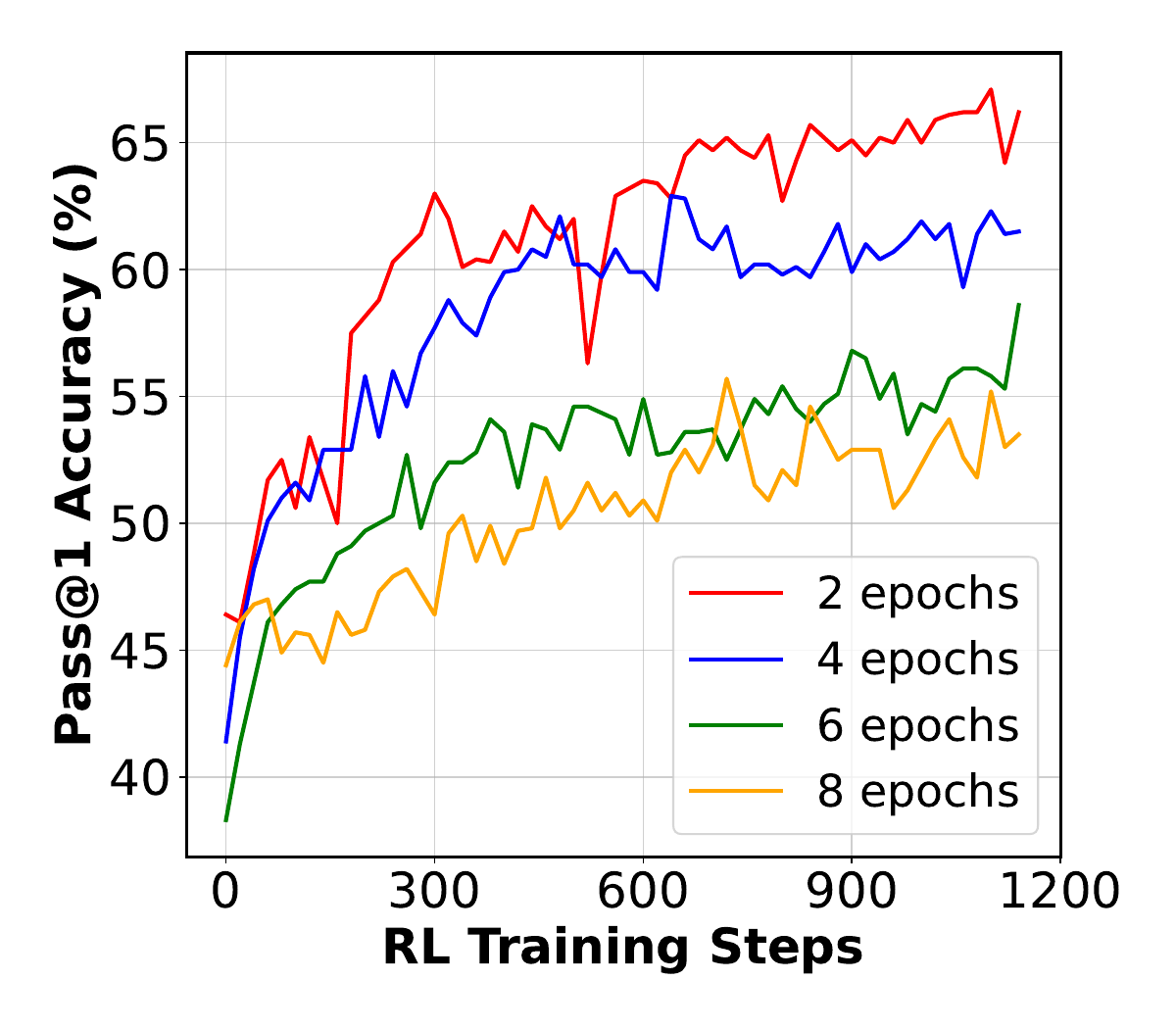}
        \caption{Performance}
        \label{fig:perform}
    \end{subfigure}
    \hfill
    \begin{subfigure}[b]{0.32\linewidth}
        \includegraphics[width=\linewidth]{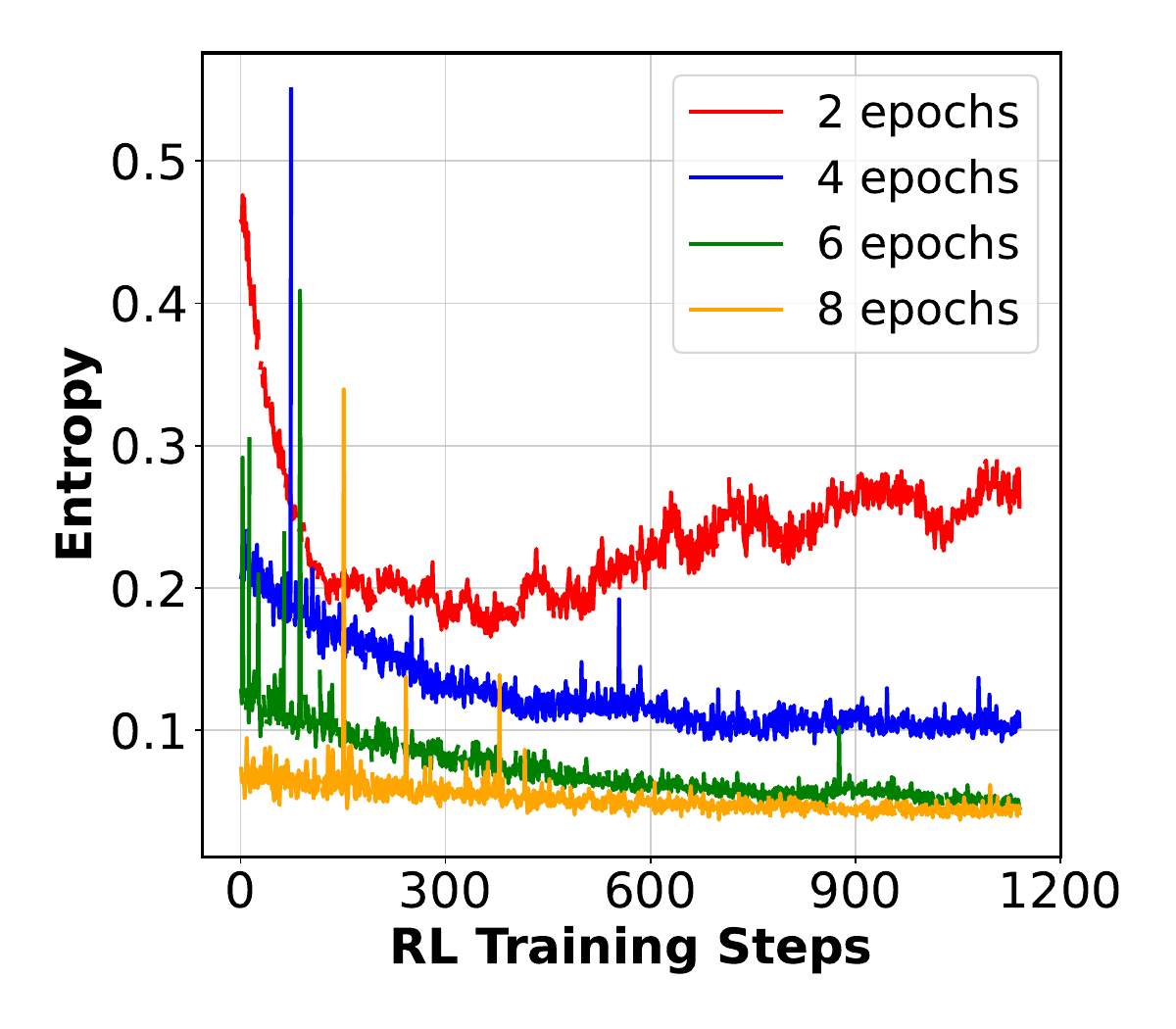}
        \caption{Entropy}
        \label{fig:entropy}
    \end{subfigure}
    \hfill
    \begin{subfigure}[b]{0.32\linewidth}
        \includegraphics[width=\linewidth]{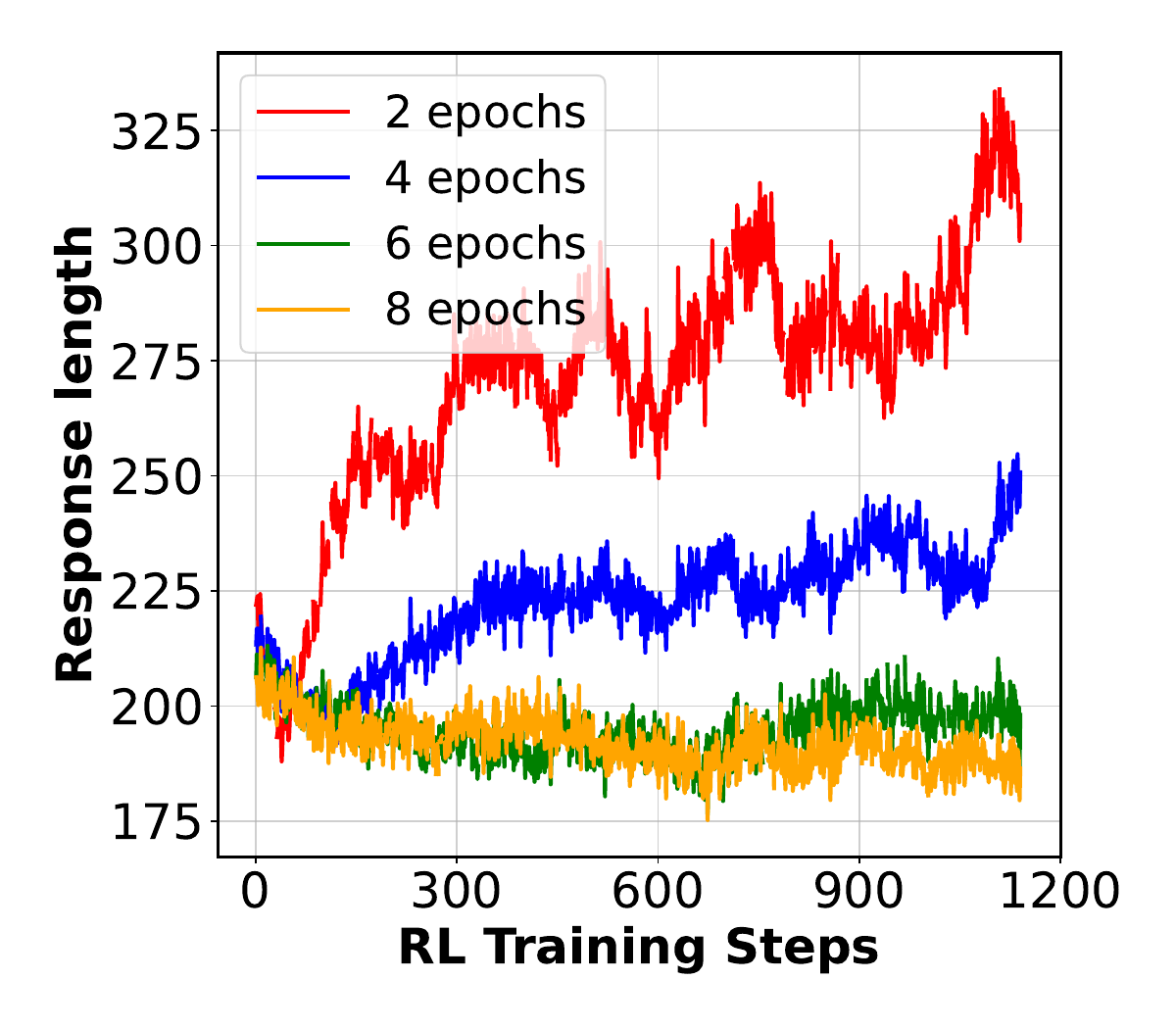}
        \caption{Response Length}
        \label{fig:length}
    \end{subfigure}
    \caption{Performance, entropy, and response length on the PGPS9K test set during RL training varying training epochs of SFT initialization.}
    \label{fig:sft_epoch_ablation}
\end{figure}

\subsubsection{Training}
\label{sec:training}

\textbf{Moderate SFT Preserves Reasoning Potential.} Previous studies in natural language reasoning~\citep{li2025preserving,zeng2025simplerl} suggest that excessive SFT may compromise response diversity, consequently limiting the model's reasoning capacity establishment in RL training. Our experiment results also verify this point. We perform an ablation study examining varying durations of cold start training (epochs) and their impact on subsequent RL performance. As illustrated in Figure~\ref{fig:sft_epoch_ablation}, intensive fine-tuning (8 epochs, yellow color line) yields diminishing performance, showing sluggish performance gains during RL post-training. While moderate fine-tuning (2 epochs, red color line) achieves substantially better RL training efficiency. Additional entropy and response length measurements during RL training further confirm that the moderate cold start preserves response diversity, creating a more exploratory policy space that ultimately unlocks greater reasoning potential during RL post-training.

\begin{figure}[t]
    \centering
    \includegraphics[width=0.85\linewidth]{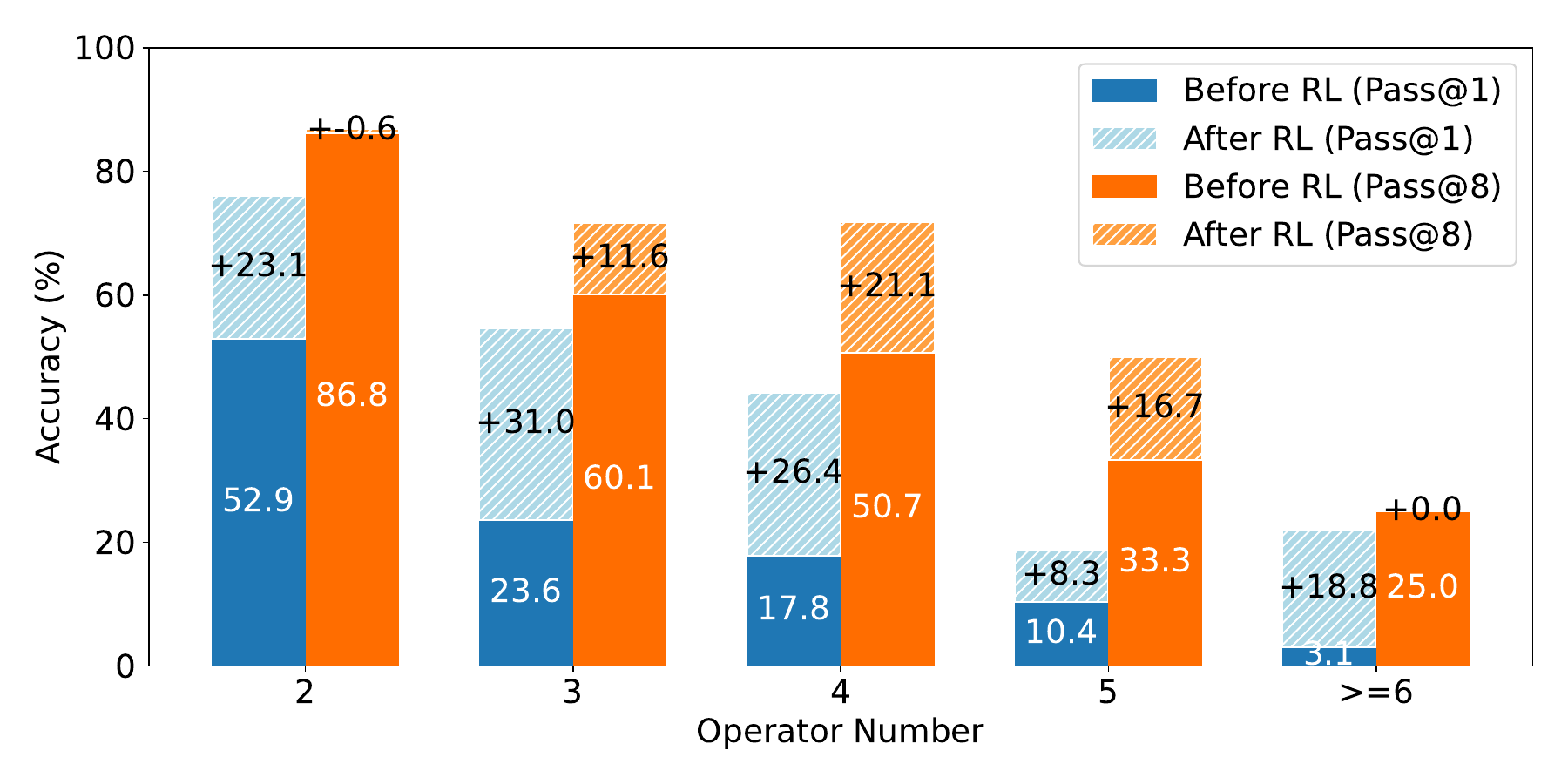}
    \caption{Pass@K accuracy on PGPS9K before/after RL training across different operator numbers.}
    \label{fig:sft_rl_operator_number}
\end{figure}

\textbf{RL Training Enhances Test-Time Scaling for Medium-Difficulty Problems but Saturates on Easy/Hard Ones.} To investigate how RL training impacts problem-solving capability, we stratify the PGPS9k test problems into five levels based on operator count (2, 3, 4, 5, and $\geq$6), and report the Pass@K performance across five difficulty levels before and after RL training. As indicated in ~\cref{fig:sft_rl_operator_number}, we could observe that: 1) Model performance after RL monotonically decreases with complexity increasing, from 86.2\% Pass@8 (operator count = 2) to 25.0\% Pass@8 (operator count $\geq$ 6), confirming operator count as a reliable proxy for solver capability boundaries.
2) RL selectively enhances Pass@8 performance, delivering significant gains only for medium-difficulty problems (operator=3-5), with improvements of +11.6\% (3 ops), +21.1\% (4 ops), and +16.7\% (5 ops) respectively. Neither easy (operator=2) nor harder problems (operator $\ge$ 6) show little significant improvement, suggesting RL primarily optimizes tasks where the base model has learnable but suboptimal strategies.

\section{Conclusion}

In this paper, we introduce a new hybrid reasoning paradigm for solving geometry problems, combining natural language with formal reasoning steps to leverage the complementary strengths of both approaches. To facilitate this framework, we curate a new 11k-sample dataset featuring formal-integrated CoT reasoning, including auto-formalization and natural-formal interleaved reasoning trajectories. Using this dataset, we investigate post-training procedures such as SFT and RL to train models in hybrid reasoning. Our results demonstrate that GF-Reasoner, our trained model integrating formal language with CoT reasoning, yields significant improvements in both geometric problem-solving performance and token efficiency.  We further provide a comprehensive analysis of critical design choices (e.g., reasoning paradigm, dataset composition, and training strategies), offering valuable insights for future research in this direction.

While our geometric formal language paradigm demonstrates significant improvements in reducing computational and reasoning errors, it is not perfect with limitations. First, the current framework cannot handle problems requiring diagrammatic constructions (e.g., adding auxiliary lines), which restricts its applicability to certain classes of geometry problems. Second, our method cannot perform long-horizon reasoning with self-reflection as in~\citep{huang2025vision} yet. We believe empowering our framework with this feature would further enhance the performance. Additionally, our approach receives limited reward supervision during training, in contrast to the rich executor feedback and multi-turn interaction mechanisms used by \citep{li2025cort}. Exploring these enhancements is a promising direction for future work.

\bibliographystyle{named}
\bibliography{reference.bib}

\input{appendix/appendix}

\end{document}

%% file: math_commands/header.tex
\usepackage{url}            
\usepackage{booktabs}       
\usepackage{multirow}    
\usepackage{amsfonts}       
\usepackage{nicefrac}       
\usepackage{microtype}      
\usepackage{natbib}
\usepackage{enumerate}
\usepackage{hhline}
\usepackage{makecell}
\usepackage{pifont}

\usepackage{graphicx} 
\usepackage{caption}
\usepackage{subcaption}
\usepackage{amsmath}
\usepackage{amsthm}
\usepackage{amssymb}
\usepackage{tikz}
\usepackage{xcolor}
\usetikzlibrary{arrows}

\allowdisplaybreaks

\usepackage{mathrsfs}

\usepackage{algorithm}
\usepackage{algpseudocode}
\usepackage{hyperref}
\usepackage{bm}

\allowdisplaybreaks

\newcolumntype{C}[1]{>{\centering\arraybackslash}m{#1}}

\usepackage[capitalize]{cleveref}
\crefname{thm}{Theorem}{Theorems}
\crefname{lem}{Lemma}{Lemmas}
\crefname{cor}{Corollary}{Corollaries}
\crefname{prop}{Proposition}{Propositions}
\crefname{asmp}{Assumption}{Assumptions}
\crefname{defn}{Definition}{Definitions}
\crefname{oracle}{Oracle}{Oracles}
\crefname{fact}{Fact}{Facts}
\crefname{conj}{Conjecture}{Conjectures}
\crefname{rem}{Remark}{Remarks}
\crefname{example}{Example}{Examples}
\crefname{condition}{Condition}{Conditions}
\crefname{exercise}{Exercise}{Exercises}
\crefname{algorithm}{Algorithm}{Algorithms}
\crefname{table}{Table}{Tables}
\crefname{figure}{Figure}{Figures}
\crefname{section}{Section}{Sections}
\crefname{subsection}{Section}{Sections}
\crefname{appendix}{Appendix}{Appendices}
\crefname{mess}{Message}{Messages}
\crefname{claim}{Claim}{Claims}
\crefname{question}{Question}{Questions}
\crefname{case}{Case}{Cases}

\definecolor{red}{rgb}{1, 0, 0}

\definecolor{green}{rgb}{0, 1, 0}

\definecolor{blue}{rgb}{0, 0, 1}

\definecolor{orange}{rgb}{1, 0.4, 0.0}

%% file: appendix/appendix.tex
\appendix 

\input{appendix/prompts}
\input{appendix/experiment_results}

\input{appendix/experiment_details}

%% file: appendix/prompts.tex
\clearpage
\section{Dataset Details}
\label{appendix:prompt}

We provide forward and backward synthetic prompts for generating Formal-integrated CoT data samples in~\cref{appendix:forward} and~\cref{appendix:backward}.~\cref{appendix:error_cases_prompting} provides a case study on directly prompting the base model (Qwen-VL-2.5-7B) with human demonstrations. The results show that the base model lacks prior knowledge of geometric formal language. Finally, we showcase Formal-integrated CoT examples from our dataset in~\cref{appendix:dataset_examples}.

\subsection{Forward Synthetic Prompt}
\label{appendix:forward}

\begin{tcolorbox}[colback=gray!20, breakable, colframe=gray, title=Forward Synthesis Prompt]

You are a geometry problem expert. You have access to a solver with the following formal language. The following prompt teaches you how to use this language through examples and explanations.

\section*{Formal Language Specification}

\subsection*{Operands}
This language uses three types of variables:
\begin{itemize}
    \item Input values (\texttt{N0}, \texttt{N1}, \texttt{N2}, etc.): These represent given measurements, lengths, angles, or other known values in a problem. The numbering must start from \texttt{N0} and increment by 1.
    \item Calculated values (\texttt{V0}, \texttt{V1}, \texttt{V2}, etc.): These store intermediate calculations or final results. The numbering must start from \texttt{V0} and increment by 1.
    \item Constants (\texttt{C0.5}, \texttt{C2}, \texttt{C3}, \texttt{C90}, etc.): These represent common numerical values like 0.5, 2, 3, 90, etc.
\end{itemize}

\subsection*{Operators}
The language includes 34 operators that represent geometric relations and calculations:

\subsubsection*{Basic Mathematical Operations}
\begin{itemize}
    \item \texttt{Get}: Retrieves the numerical value of a variable (e.g., \texttt{Get V0} returns the value stored in \texttt{V0})
    \item \texttt{Sum}: Performs addition of multiple terms (e.g., \texttt{Sum a b c d} represents $a+b+c=d$)
    \item \texttt{Multiple}: Performs multiplication of multiple terms (e.g., \texttt{Multiple a b c d} represents $a \times b \times c = d$)
    \item \texttt{Equal}: Sets two expressions equal (e.g., \texttt{Equal a b} represents $a = b$)
\end{itemize}

\subsubsection*{Triangle Operations}
\begin{itemize}
    \item \texttt{Gougu}: Implements the Pythagorean theorem (e.g., \texttt{Gougu a b c} represents $a^2 + b^2 = c^2$)
    \begin{itemize}
        \item 'a' is the first leg of the right-angled triangle
        \item 'b' is the second leg of the right-angled triangle
        \item 'c' is the hypotenuse opposite the right angle
    \end{itemize}
    $\cdots$
\end{itemize}

\subsubsection*{Trigonometric Operations}
\begin{itemize}
    \item \texttt{Gsin}: Implements sine relation (e.g., \texttt{Gsin a b c} represents $\sin(c) = \frac{a}{b}$)
    \begin{itemize}
        \item 'a' is the opposite side length
        \item 'b' is the hypotenuse length
        \item 'c' is the angle in degrees
    \end{itemize}
    $\cdots$
\end{itemize}

\subsubsection*{Quadrilateral Operations}
\begin{itemize}
    \item \texttt{Para\_Area}: Calculates parallelogram area (e.g., \texttt{Para\_Area a b c} represents $a \times b = c$)
    \begin{itemize}
        \item 'a' is the base length
        \item 'b' is the height perpendicular to the base
        \item 'c' is the resulting area
    \end{itemize}
    $\cdots$
\end{itemize}

\subsubsection*{Circle Operations}
\begin{itemize}
    \item \texttt{Circle\_R\_Circum}: Calculates circle circumference from radius (e.g., \texttt{Circle\_R\_Circum a b} represents $2\pi \times a = b$)
    \begin{itemize}
        \item 'a' is the radius length
        \item 'b' is the resulting circumference
        \item Optional format: \texttt{Circle\_R\_Circum a b c} represents $2\pi \times a \times \frac{b}{360} = c$, where 'b' is the central angle in degrees and 'c' is the arc length.
    \end{itemize}
    $\cdots$
\end{itemize}

\subsubsection*{Other Geometric Relations}
\begin{itemize}
    \item \texttt{Geo\_Mean}: Calculates geometric mean (e.g., \texttt{Geo\_Mean a b c} represents $\sqrt{a \times b} = c$)
    \begin{itemize}
        \item 'a' is the first value
        \item 'b' is the second value
        \item 'c' is the resulting geometric mean
        \item This operator has numerous applications in geometric problems:
        \begin{itemize}
            \item Tangent-Secant Power Theorem: \texttt{Geo\_Mean PA PB PT} $\rightarrow$ PT is the tangent length when a point P outside the circle has a secant with segments PA and PB, where PT = $\sqrt{PA \times PB}$.
            \item Altitude Rule (Right Triangle Projection): \texttt{Geo\_Mean p q h} $\rightarrow$ In a right triangle, the altitude h to the hypotenuse is the geometric mean of the two segments p and q it creates on the hypotenuse, where h = $\sqrt{p \times q}$.
            \item Leg Projection Rule: \texttt{Geo\_Mean p c a} $\rightarrow$ A leg a of a right triangle is the geometric mean of its projection p on the hypotenuse and the hypotenuse c itself, where a = $\sqrt{p \times c}$.
        \end{itemize}
    \end{itemize}
    $\cdots$
\end{itemize}

\section*{Solution Format}
\begin{itemize}
    \item Given the problem, you should first understand it and analyze the image to identify important information that can lead to the solution.
    \item Reason step-by-step using the above Operators and Operands to arrive at the final solution. You must explicitly state the Operator and Operand you used in your reasoning steps. You may use multiple sequences of Operators and Operands.
    \item You do not need to calculate the numerical value in your reasoning steps since the solver with your formal language answer can do this.
    \item Put your final formal language solution and parameter values in \texttt{\textbackslash boxed\{\}} separately, both on a single line. Ensure that the content inside \texttt{\textbackslash boxed\{\}} is space-delimited and compatible with Python's \texttt{eval()} for direct parsing by the solver.
\end{itemize}

\section*{Examples}

\subsubsection*{Example 1}
\textbf{Problem:} Find the area of the figure.

\textbf{Image:}
\begin{center}
    \includegraphics[width=0.3\textwidth]{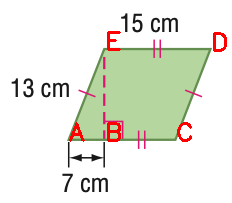}
\end{center}

\textbf{Step-by-Step Thinking:}
Looking at the image, I see a parallelogram AEDC with various measurements. From the image and problem statement: - $AC \perp EB$ at point $B$ - $ED = AC = 15$ cm - $AE = DC = 13$ cm - $AB = 7$ cm

I need to find the area of the parallelogram AEDC. Let me solve this by using geometric principles and operations from the formal language:

\begin{enumerate}
    \item Identify the important information:
    \begin{itemize}
        \item $AC \perp EB$ at point $B$
        \item $ED = AC = 15$ cm
        \item $AE = DC = 13$ cm
        \item $AB = 7$ cm
    \end{itemize}
    
    \item Understanding the Solution Path:
    \begin{itemize}
        \item To find the area of a parallelogram, I need the base and height.
        \item I can use $ED = 15$ cm as the base.
        \item For the height, I need the perpendicular distance from $ED$ to $AC$.
        \item The problem states $AC \perp EB$ at $B$, which means $AC$ is perpendicular to $EB$.
        \item I need to find the height $V0$ using the Pythagorean theorem with $AB$ and the distance from $A$ to the line $ED$.
        \item Given $AB = 7$ cm and the distance from $A$ to $E$ is $13$ cm, I can use the Pythagorean theorem to find the height.
        \item Once I have the height $V0$, I can calculate the area as base $\times$ height.
    \end{itemize}
    
    \item Step-by-Step Reasoning Using Operators and Operands:
    \begin{itemize}
        \item First, let $ED = AC = \texttt{N0}$, $AE = DC = \texttt{N1}$, $AB = \texttt{N2}$.
        \item Then, I use the Pythagorean theorem to find the height \texttt{V0}:
        \begin{center}
            \texttt{Gougu V0 N2 N1}
        \end{center}
        This solves for \texttt{V0} where $\texttt{V0}^2 + \texttt{N2}^2 = \texttt{N1}^2$
        
        \item Then, I calculate the area of the parallelogram \texttt{V1} using base $\times$ height:
        \begin{center}
            \texttt{Para\_Area N0 V0 V1}
        \end{center}
        This computes $\texttt{V1} = \texttt{N0} \times \texttt{V0}$
        
        \item Finally, I get the result:
        \begin{center}
            \texttt{Get V1}
        \end{center}
    \end{itemize}
    
    \item Formal Language Solution:
    \begin{center}
        \boxed{\texttt{Gougu V0 N2 N1 Para\_Area N0 V0 V1 Get V1}}
    \end{center}
    
    with parameter values:
    \begin{center}
        \boxed{\texttt{N0=15, N1=13, N2=7}}
    \end{center}
\end{enumerate}

\subsubsection*{Example 2}
$\cdots$
\section*{Your Task}
\textbf{Problem:} \{\{Target problem\}\}

\textbf{Image:} \{\{Image diagram of the target problem\}\}

\textbf{Step-by-Step Thinking:}
\end{tcolorbox}

\clearpage

\subsection{Backward Synthesis Prompt}
\label{appendix:backward}

\begin{tcolorbox}[colback=gray!20, breakable, colframe=gray, title=Customized Backward Synthesis Prompt for \texttt{Chord2\_Ang\_v0}]

You are a geometry problem expert. You have access to a solver with the following formal language. The following prompt teaches you how to use this language through examples and explanations.

\section*{Formal Language Specification}

$\cdots$

$\langle \text{Same with the forward synthesis prompt} \rangle$
 
$\cdots$

\section*{Examples}

\subsubsection*{Example 1}
\textbf{Problem:} Find $m\angle F$.

\textbf{Image:}
\begin{center}
    \includegraphics[width=0.3\textwidth]{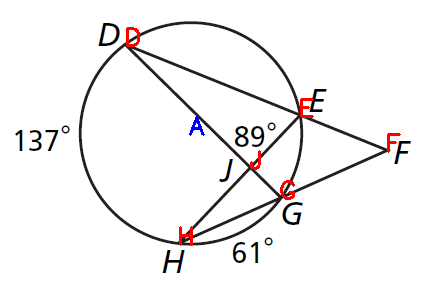}
\end{center}

\textcolor{blue}{\textbf{Solution:} \texttt{Sum V0 N0 C180 Chord2\_Ang V0 V1 N1 TanSec\_Ang V2 V1 N1 Get V2}}

\paragraph{Step-by-Step Thinking:}
Looking at the circle diagram, I can see several points ($D$, $E$, $F$, $G$, $H$, $J$) with some angles marked:
\begin{itemize}
    \item $m\angle DJE = 89^\circ$
    \item $m\angle HG$ (arc $HG$) $= 61^\circ$
    \item $m\angle DH$ (arc $DH$) $= 137^\circ$
    \item I need to find $m\angle F$
\end{itemize}

The task is to find angle $F$. Let me solve this by using geometric principles and operations from the formal language:

1. Identify the Given Information:
\begin{itemize}
    \item $m\angle DJE = 89^\circ$
    \item $m\angle HG$ (arc $HG$) $= 61^\circ$
    \item $m\angle DH$ (arc $DH$) $= 137^\circ$
    \item Point $J$ is inside the circle
    \item Point $F$ is outside the circle
    \item Need to find $m\angle F$
\end{itemize}

2. Understanding the Solution Path:
\begin{itemize}
    \item First, I notice that points $D$, $H$, $G$, and $E$ lie on the circle. Point $J$ is inside the circle, while point $F$ is outside the circle.
    \item Since $F$ is outside the circle with secant lines $FD$ and $FH$, I can use the \texttt{TanSec\_Ang} theorem
    \item To apply this theorem, I need to find the intercepted arcs
    \item To find one of the arcs, I'll use the \texttt{Chord2\_Ang} operator and properties of angles inside a circle
\end{itemize}

3. Step-by-Step Reasoning Using Operators and Operands:
\begin{itemize}
    \item I'll first calculate the supplementary angle to $\angle DJE$, let $\angle DJE = N0$ and $\angle EJG = V0$:\\
      \texttt{Sum V0 N0 C180}\\
      (This calculates $V0 = 180^\circ - N0 = 180^\circ - 89^\circ = 91^\circ$)
      
    \item Since $J$ is a point inside the circle, I can use the \texttt{Chord2\_Ang} theorem to calculate the arc $EG$, let arc $DH = N1$ and arc $EG = V1$:\\
      \texttt{Chord2\_Ang V0 V1 N1}\\
      (This relationship comes from the inscribed angle theorem for intersecting chords)
      
    \item Finally, since $F$ is outside the circle with secant lines to $D$ and $E$, I can use the \texttt{TanSec\_Ang} theorem to find $\angle F$, let $\angle F = V2$:\\
      \texttt{TanSec\_Ang V2 V1 N1}\\
      (This calculates $V2 = (N1 - V1)/2$, using the relationship between the angle formed by secant lines from an external point and the intercepted arcs)
      
    \item The final answer is obtained by:\\
      \texttt{Get V2}
\end{itemize}

4. Formal Language Solution:

\fbox{\texttt{Sum V0 N0 C180 Chord2\_Ang V0 V1 N1 TanSec\_Ang V2 V1 N1 Get V2}}

with parameter values:

\fbox{\texttt{N0=89, N1=137}}

\subsubsection*{Example 2}
$\cdots$

\section*{Your Task}
\textbf{Problem:} \{\{Target problem\}\}

\textbf{Image:} \{\{Image diagram of the target problem\}\}

\textcolor{blue}{\textbf{Solution:} \{\{Program solution of the target problem\}\}}

\textbf{Step-by-Step Thinking:}

\end{tcolorbox}

\subsection{Error Cases of Forward Prompting}
\label{appendix:error_cases_prompting}

We prompt the LVLM using manually crafted formal-integrated CoT demonstrations paired with target questions. However, we find that this method often fails. As demonstrated in the following cases, the model frequently reverts to natural language instead of maintaining formal language usage during critical reasoning steps, and it often employs operators undefined in our formal language system. These observations suggest that the model inherently lacks knowledge of geometric formal language.

\begin{tcolorbox}[colback=gray!20, breakable, colframe=gray, title=Case 1: Use Natural Language Rather Than Formal Language (Generated by Qwen-VL-2.5-7B)]

\textbf{Problem:}

line m is the perpendicular bisector of X Z , W Z = 14.9 . line m $\perp$ X Z on Y , X Y = Y Z , find W X .

\textbf{Image:}

\begin{center}
    \includegraphics[width=0.3\textwidth]{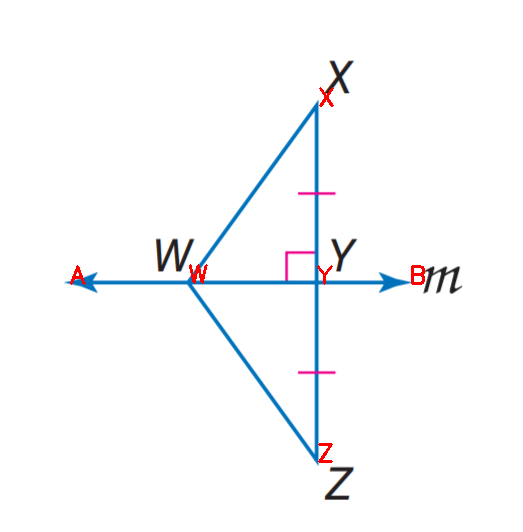}
\end{center}

\textbf{Output:}

Identify the given information:
\begin{itemize}
    \item Line \( m \) is the perpendicular bisector of \( XZ \).
    \item \( WZ = 14.9 \).
    \item \( m \perp XZ \) at \( Y \).
    \item \( XY = YZ \).  
\end{itemize}

$\cdots$\\

Apply the Pythagorean theorem:
\[
WZ^2 = WX^2 + XY^2
\]
\begin{itemize}
    \item Since \( XY = YZ \), \( XY = \frac{14.9}{2} = 7.45 \).
\end{itemize}

Substitute the known values:
\[
14.9^2 = a^2 + 7.45^2
\]

$\cdots$ \\

Final calculation:

$a = \sqrt{166.5075} \approx 12.9$

Formal Language Solution:
\boxed{\texttt{W X = sqrt(166.5075)}}
with parameter values:
\boxed{\texttt{a=12.9}}\\

\textbf{Error:}

The model directly use natural language to solver the problem without following the instruction prompt.

\end{tcolorbox}

\begin{tcolorbox}[colback=gray!20, breakable, colframe=gray, title=Case 2: Use Undefined Operators (Generated by Qwen-VL-2.5-7B)]

\textbf{Problem:}

C D = 9 , C B = C D = x+3 , A B = 7y-2 , D A = 30 , find the length of D B .

\textbf{Image:}

\begin{center}
    \includegraphics[width=0.3\textwidth]{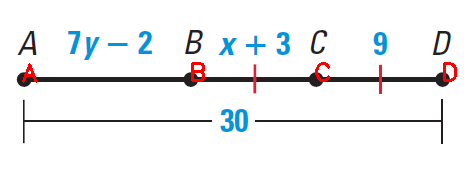}
\end{center}

\textbf{Output:}

$\cdots$ \\

3. Step-by-Step Reasoning Using Operators and Operands:\\
  - First, solve for x:\\
    \texttt{Equal x + 3 9}\\
    \texttt{Solve x}\\
    Let \texttt{x = N0}\\

  - Then, calculate C B:\\
    \texttt{Equal N0 + 3 V0}\\
    \texttt{Set V0 = 9}\\

  - Next, calculate D B:\\
    \texttt{Equal D A V1 D C V2 C B V3}\\
    \texttt{Equal 30 V1 9 V2 9 V3}\\
   \texttt{Solve V3}\\

4. Formal Language Solution:\\
\boxed{\texttt{Equal x + 3 9 Solve x Equal N0 + 3 V0 Set V0=9 Equal 30 V1 9 V2 9 V3}}
\boxed{\texttt{Solve V3}}

with parameter values:\\ 
\boxed{\texttt{x=6 V0=9 V3=12}}\\

\textbf{Error:}

\texttt{Solve} and \texttt{Set} are undefined operators in the formal language.

\end{tcolorbox}

\subsection{Examples of Formal-Integrated CoT Dataset}
\label{appendix:dataset_examples}

We provide some examples in our synthetic formal-integrated CoT dataset below.

\begin{tcolorbox}[colback=gray!20, breakable, colframe=gray, title=Formal-Integrated CoT Example1]

\textbf{Problem:}

assume that all polygons that appear to be regular are regular . $G C = 8$ , find the area of the shaded region .

\textbf{Image:}

\begin{center}
    \includegraphics[width=0.3\textwidth]{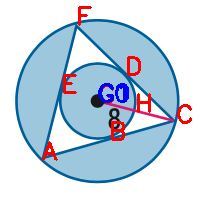}
\end{center}

\textbf{Output:}

First, label the given length \texttt{N0 = 8} (GC is the radius of the outer circle).

\begin{enumerate}
    \item Compute the area of the outer circle (radius \texttt{N0}):
    \begin{itemize}
        \item \texttt{Circle\_R\_Area N0 V0}
        \item $V0 = \pi \cdot \texttt{N0}^2$
    \end{itemize}
    
    \item Compute the area of the regular triangle inscribed in that outer circle. A regular 3-gon with circumradius \texttt{N0} has area:
    \begin{itemize}
        \item \texttt{RNgon\_L\_Area C3 N0 V1}
        \item $V1 = \frac{3 \cdot \texttt{N0}^2 \cdot \sin(120^\circ)}{2}$
    \end{itemize}
    
    \item Find the apothem (inradius) \texttt{V2} of that same triangle by using its area \texttt{V1}:
    \begin{itemize}
        \item \texttt{RNgon\_H\_Area C3 V2 V1}
        \item $V1 = 3 \cdot \texttt{V2}^2 \cdot \tan(60^\circ) \Rightarrow \texttt{V2}$ is the apothem
    \end{itemize}
    
    \item Compute the area of the inner circle (incircle of the triangle) of radius \texttt{V2}:
    \begin{itemize}
        \item \texttt{Circle\_R\_Area V2 V3}
        \item $V3 = \pi \cdot \texttt{V2}^2$
    \end{itemize}
    
    \item The shaded region consists of (outer circle minus the triangle) plus the inner circle. First form the difference outer - triangle:
    \begin{itemize}
        \item \texttt{Sum V1 V4 V0}
        \item $V1 + V4 = V0 \Rightarrow V4 = V0 - V1$
    \end{itemize}
    
    \item Then add in the inner circle:
    \begin{itemize}
        \item \texttt{Sum V3 V4 V5}
        \item $V3 + V4 = V5$
    \end{itemize}
    
    \item Finally output the total shaded area \texttt{V5}:
    \begin{itemize}
        \item \texttt{Get V5}
    \end{itemize}
\end{enumerate}

Putting it all together:

\begin{center}
    \boxed{\texttt{Circle\_R\_Area N0 V0 RNgon\_L\_Area C3 N0 V1 RNgon\_H\_Area C3 V2 V1}}
    \boxed{\texttt{Circle\_R\_Area V2 V3 Sum V1 V4 V0 Sum V3 V4 V5 Get V5}}
\end{center}

with 

\begin{center}
    \boxed{\texttt{N0=8}}
\end{center}

\end{tcolorbox}

\begin{tcolorbox}[colback=gray!20, breakable, colframe=gray, title=Formal-Integrated CoT Example2]

\textbf{Problem:}

$QP \parallel RT$, $QR = 2$, $SR = PT = 3$, $PQ = x$, $m\angle TQP = m\angle RQT$, solve for $x$.

\textbf{Image:}

\begin{center}
    \includegraphics[width=0.3\textwidth]{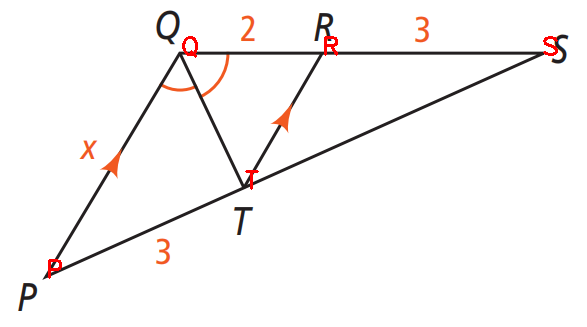}
\end{center}

\textbf{Output:}

Here is the step-by-step setup using our formal language:

Define operands:
\begin{itemize}
    \item \texttt{N0} = QR = 2  
    \item \texttt{N1} = SR = PT = 3  
    \item \texttt{N2} = PQ = x  
\end{itemize}

1. Since QP $\parallel$ RT, by similar triangles we have ST/PT = SR/QR. Let ST = \texttt{V0}:  
   \begin{itemize}
       \item \textbf{Operator}: \texttt{Proportion V0 N1 N1 N0}  
       \item This establishes $\frac{\texttt{V0}}{\texttt{N1}} = \frac{\texttt{N1}}{\texttt{N0}}$
   \end{itemize}

2. QS is the sum of QR and RS. Let QS = \texttt{V1}:  
   \begin{itemize}
       \item \textbf{Operator}: \texttt{Sum N0 N1 V1}  
       \item Calculates $\texttt{V1} = \texttt{N0} + \texttt{N1} = 2 + 3 = 5$
   \end{itemize}

3. In $\triangle$PQS, QT bisects $\angle$PQS, so PT/ST = PQ/QS:  
   \begin{itemize}
       \item \textbf{Operator}: \texttt{Proportion N1 V0 N2 V1}  
       \item Establishes $\frac{\texttt{N1}}{\texttt{V0}} = \frac{\texttt{N2}}{\texttt{V1}}$
   \end{itemize}

4. Solve for PQ (\texttt{N2}):  
   \begin{itemize}
       \item \textbf{Operator}: \texttt{Get N2}  
       \item Retrieves the value of $\texttt{N2} = x$
   \end{itemize}

Final formal-language solution and parameters:

\begin{center}
    \boxed{\texttt{Proportion V0 N1 N1 N0 Sum N0 N1 V1 Proportion N1 V0 N2 V1 Get N2}}
\end{center}

\begin{center}
    \boxed{\texttt{N0=2 N1=3 N2=x}}
\end{center}

\end{tcolorbox}

%% file: appendix/experiment_results.tex
\clearpage

\section{Additional Results}
\label{appendix:additional_results}

In~\cref{appendix:error_catogory}, we illustrate four types of errors in geometry problem solving. In~\cref{appendix:error_reduction_case}, we present illustrative typical examples demonstrating how formal-integrated reasoning reduces both computational and reasoning errors. In~\cref{appendix:larger_model}, we provide SFT initialization results on a larger base model (Qwen-2.5-VL-32B). In~\cref{appendix:compare_o4}, we provide a comparative analysis with the frontier large multimodal model OpenAI-o4-mini. In~\cref{appendix:compare_paradigm}, we provide a paradigm comparison with existing approaches.

\subsection{Categorization of Geometry Problem-Solving Errors}
\label{appendix:error_catogory}

In~\cref{sec:analysis}, we categorize geometry problem-solving errors into four types following existing works. The illustration of the four types of errors is described below:
\begin{itemize}
    \item Visual perception errors: The model failed to either extract salient features from the image or associate the textual information with its corresponding geometric shapes.
    \item Reasoning errors: The model failed to generate a valid solution, demonstrating either a lack of logical coherence and internal consistency or a failure in procedural execution by skipping necessary operations.
    \item Geometric knowledge errors: The model's error was either in misapplying a geometric concept or in failing to distinguish between the definitions of two theorems.
    \item Computation errors: The model failed to calculate the numerical results, leading to errors in calculations involving arithmetic, square roots, and proportions.
\end{itemize}

We classify the wrong solution into one of the four types of errors based on the first error in the response. To better illustrate, we provide cases for the four types of errors below. The responses are generated by Qwen-VL-2.5-7B.

\begin{center}
\begin{minipage}{0.85\linewidth} 

\begin{tcolorbox}[colback=white, colframe=DarkBlue, title=\textbf{\#1 Visual Perception Error},arc=3mm,boxrule=0.8pt]

\textbf{Problem}\\
E A parallel D B , E D = 4 , C B = 12 , A B = 5 , C D = x , find x .
\tcbline

\textbf{Image}\\
\makebox[\linewidth]{\includegraphics[width=0.3\linewidth]{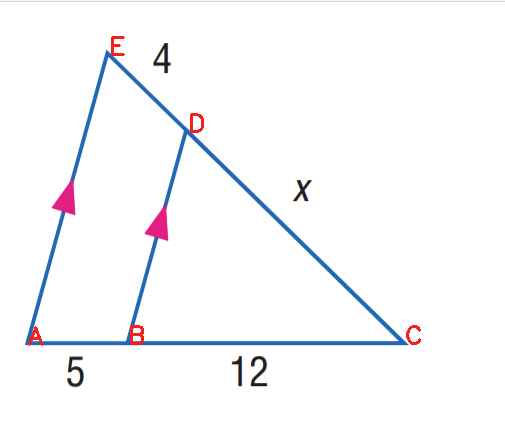}}
\tcbline

\textbf{\textcolor{red}{Model's Response}}
\vspace{10pt}\\
\begin{tabularx}{\linewidth}{@{} l L @{}}
 \textbf{Response} & ...we need to use the properties of similar triangles. Since EA \(\parallel\) DB, \textcolor{red}{\( \triangle\) AED and \( \triangle\) ABC are similar} by the AA (Angle-Angle) similarity criterion.  \\
 \textbf{Error} & $\triangle$ AED and $\triangle$ ABC are not triangle. \\
\end{tabularx}
\tcbline

\textbf{\textcolor{DarkGreen}{Correct Solution}}
\vspace{10pt}\\
\begin{tabularx}{\linewidth}{@{} l L @{}}
    \textbf{Thought} & Since line segment EA is parallel to DB, it follows that triangle CEA is similar to triangle CDB (\( \triangle\) CEA $\sim$ \( \triangle\) CDB).  \\
    \textbf{Answer} & x = 9.6 
\end{tabularx}

\end{tcolorbox}
\end{minipage}
\end{center}

\begin{center}
\begin{minipage}{0.85\linewidth} 

\begin{tcolorbox}[colback=white, colframe=DarkBlue, title=\textbf{\#2 Reasoning Error},arc=3mm,boxrule=0.8pt]

\textbf{Problem}\\
    m$\angle$AXY = 22, m$\angle$ AYX = 2x+2, and m$\angle$ AZW = 3x-23, what is m$\angle$ W?
\tcbline

\textbf{Image}\\
\makebox[\linewidth]{\includegraphics[width=0.3\linewidth]{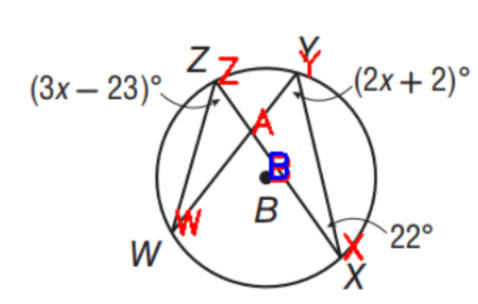}}
\tcbline

\textbf{\textcolor{red}{Model's Response}}
\vspace{10pt}\\
\begin{tabularx}{\linewidth}{@{} l L @{}}
 \textbf{Response} & ...\textcolor{red}{If we assume $x$=23} (a common assumption for such problems without additional constraints), then... \\
 \textbf{Error} & The model couldn't solve for $x$ geometrically, so it instead assumed a ``common" value of $x$=23, a step not based on the problem's logic. \\
\end{tabularx}
\tcbline

\textbf{\textcolor{DarkGreen}{Correct Solution}}
\vspace{10pt}\\
\begin{tabularx}{\linewidth}{@{} l L @{}}
    \textbf{Thought} & $\angle$W and $\angle$AXY are both subtended by the same arc ZY. According to the Inscribed Angle Theorem, $\angle$W = $\angle$AXY = 22°. \\
    \textbf{Answer} & m$\angle$W = 22°\\
\end{tabularx}

\end{tcolorbox}

\begin{tcolorbox}[colback=white, colframe=DarkBlue, title=\textbf{\#3 Geometric Knowledge Error},arc=3mm,boxrule=0.8pt]

\textbf{Problem}\\
    RQ = 6, QT = x, SR = 14, SQ = 18, and m$\angle$ QRT = m$\angle$ SRT, find x.
\tcbline

\textbf{Image}\\
\makebox[\linewidth]{\includegraphics[width=0.25\linewidth]{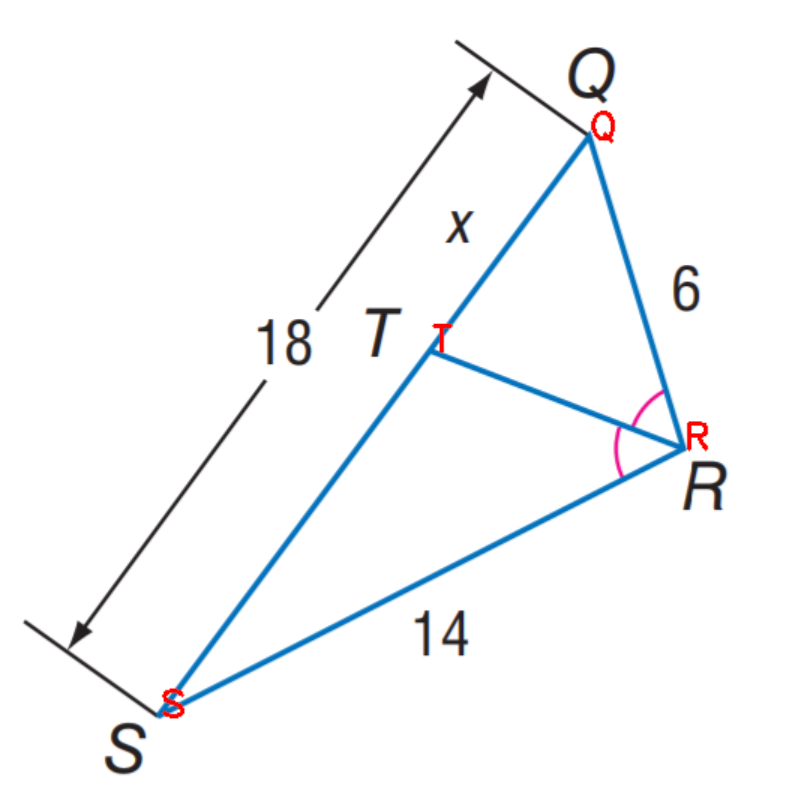}}
\tcbline

\textbf{\textcolor{red}{Model's Response}}
\vspace{10pt}\\
\begin{tabularx}{\linewidth}{@{} l L @{}}
     \textbf{Response} & ...We are given that angles ($\angle$QRT = $\angle$SRT). This implies that \textcolor{red}{ $\triangle$QTR and $\triangle$SRT are similar by the Angle-Angle (AA) similarity criterion}.... \\
     \textbf{Error} & It incorrectly concludes that $\triangle$QTR and $\triangle$SRT are similar based on having only one pair of equal angles ($\angle$QRT = $\angle$SRT). \\
\end{tabularx}
\tcbline

\textbf{\textcolor{DarkGreen}{Correct Solution}}
\vspace{10pt}\\
\begin{tabularx}{\linewidth}{@{} l L @{}}
        \textbf{Thought} & Since $\angle$ QRT = $\angle$ SRT, the segment RT bisects the $\angle$ QSR in $\triangle$ QSR. According to the Angle Bisector Theorem, QR/QT = RS/TS.  \\
        \textbf{Answer} & x = 5.4 \\
\end{tabularx}
\end{tcolorbox}

\end{minipage}
\end{center}

\begin{center}
\begin{minipage}{0.85\linewidth} 

\begin{tcolorbox}[colback=white, colframe=DarkBlue, title=\textbf{\#4 Computation Error},arc=3mm,boxrule=0.8pt]

\textbf{Problem}\\
    angle CDA = 70 , angle BAD = 110 , angle ABC = 3x-5 , angle DCB = 4x+10 , what is angle C ?
\tcbline

\textbf{Image}\\
\makebox[\linewidth]{\includegraphics[width=0.3\linewidth]{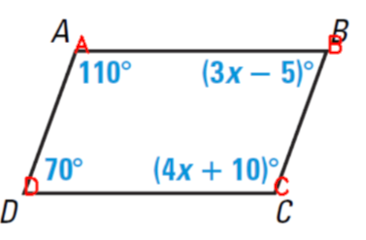}}
\tcbline

\textbf{\textcolor{red}{Model's Response}}
\vspace{10pt}\\
\begin{tabularx}{\linewidth}{@{} l L @{}}
     \textbf{Response} & To ... angle CDA + angle BAD + angle ABC + angle DCB = 360°. Substitute the given values: 70° + 110° + (3x - 5) + (4x + 10) = 360° ...\textcolor{red}{Simplify the equation: 175° + 7x = 360°}... \\
     \textbf{Error} & Calculation mistake: 70° + 110° + (3x - 5) + (4x + 10) = 360 was incorrectly simplified as 175° + 7x = 360°. \\
\end{tabularx}
\tcbline

\textbf{\textcolor{DarkGreen}{Correct Solution}}
\vspace{10pt}\\
\begin{tabularx}{\linewidth}{@{} l L @{}}
    \textbf{Thought} & The sum of the interior angles of a quadrilateral is 360°. Thus, 70 + 110 + (3x - 5) + (4x + 10) = 360 $\implies$ 185 + 7x = 360 $\implies$ x = 25 \\
    \textbf{Answer} & m$\angle$ C = 4x + 10 = 110 \\
\end{tabularx}

\end{tcolorbox}

\end{minipage}
\end{center}

\subsection{Case Study on Error Reduction}
\label{appendix:error_reduction_case}

We train a baseline model to solve geometry problems using natural language reasoning. This baseline model is trained with RL using the same base model and training dataset as our approach. The reward function is computed by comparing the numerical solution extracted from the model's response against the ground-truth answer. Below, we demonstrate how our formal-integrated reasoning model reduces computation and reasoning errors compared to this natural language baseline.


\begin{tcolorbox}[colback=white, colframe=DarkBlue, title=\textbf{Computation Error Case},arc=3mm,boxrule=0.8pt,height=5cm]
    \begin{minipage}{0.7\linewidth}
        \textbf{Question:}\\
        ED = 7, BE = EG = FG = CF = AC = AB, (m$\angle$ ABE = m$\angle$ GEB = m$\angle$ EGF = m$\angle$ GFC = m$\angle$ FCA = m$\angle$ CAB) , find the area of the shaded region.
    \end{minipage}%
    \hspace{10mm} 
    \begin{minipage}{0.2\linewidth}
        \centering
        \includegraphics[width=\linewidth,height=2cm]{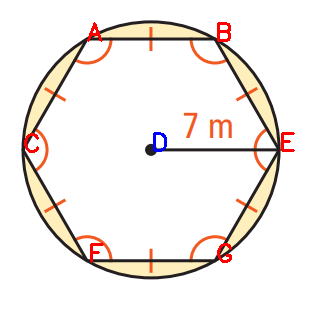} 
    \end{minipage}
\end{tcolorbox}

\begin{tcolorbox}[colback=white, title=\textbf{Natural Language Solution}, arc=3mm, boxrule=0.8pt, left=4mm, right=4mm, top=1.5mm, bottom=1.5mm, height=12cm]
    \textbf{Model's Response:}

    To find the area of the shaded region, we first need to understand that the given hexagon is inscribed in a circle with radius \(ED = 7\) (i.e., \(r = 7\)). Since the hexagon is regular, each internal angle is \(120^\circ\) (since \(360^\circ/6 = 60^\circ\), and each exterior angle is \(60^\circ\)).
    
    The area \(A_{\text{hexagon}}\) of a regular hexagon inscribed in a circle of radius \(r\) is given by:
    
    \[
    A_{\text{hexagon}} = 6 \cdot \text{Area of one equilateral triangle with side length } r
    \]
    
    The area \(A_{\text{triangle}}\) of an equilateral triangle with side length \(r\) (here \(r = 7\)) is:
    
    $
    \text{Area of equilateral triangle} = \frac{\sqrt{3}}{4} \cdot r^2
    $
    
    Substitute \(r = 7\):

    $
    \text{Area of one equilateral triangle} = \textcolor{red}{\frac{49\sqrt{3}}{4} \approx \frac{49 \cdot 1.732}{4}} \approx \frac{84.784}{4} \approx 21.196
    $
    
    Hence, area of hexagon \(A_{\text{hexagon}}\) :
    
    $
    A_{\text{hexagon}} = 6 \cdot 21.196 \approx 127.176
    $
    
    Next, area \(A_{\text{circle}}\) :

    $
    A_{\text{circle}} = \pi \cdot r^2 = \pi \cdot 7^2 = 49\pi \approx 49 \cdot 3.14159 \approx 153.94
    $
    
    Shaded area \(A_{\text{shaded}}\) :

    $
    A_{\text{shaded}} \approx 153.94 - 127.176 \approx 26.764
    $
    
    Hence, the area of the shaded region is approximately:
    
    \[
    \boxed{26.764}
    \]
    \tcbline
    
    \textbf{Answer:}\\
    26.764 (Correct: 26.632) \textcolor{red}{$\boldsymbol{\times}$} 
    
    \vspace{5pt}
    \textbf{Error:} \\
    Premature use of approximations (e.g., \(\sqrt{3} \approx 1.732\)) led to a precision error in the final result. 
\end{tcolorbox}

\begin{tcolorbox}[colback=white, colframe=DarkGreen, title=\textbf{Formal-Integrated Solution}, arc=3mm,
boxrule=0.8pt,left=4mm, right=4mm, top=1.5mm, bottom=1.5mm,height=10.5cm]
    \textbf{Model's Response:}
    
    Step-by-step reasoning: \\
    1. Let the circle’s radius (and the circumradius of the hexagon) be\\
    \( N_0 = DE = 7 \).\\
    2. Compute the area of the circle with radius \( N_0 \).\\
    Use \texttt{Circle\_R\_Area} to get the circle’s area:\\ 
    \texttt{Circle\_R\_Area} \( N_0 \) \( V_0 \) 	$\rightarrow$ \( V_0 = \pi \cdot N_0^2 \)\\
    3. Compute the area of the regular hexagon with circumradius \( N_0 \).\\
    Use \texttt{RNgon\_L\_Area} for a regular 6‐gon of circumradius \( N_0 \):\\
    \texttt{RNgon\_L\_Area} C6 \( N_0 \) \( V_1 \) 	$\rightarrow$ \( V_1 \) = area of the hexagon\\
    4. The shaded region = circle area minus hexagon area.\\
    So we take the circle area \( V_0 \) minus the hexagon area \( V_1 \).\\
    Use \texttt{Sum} to express\\
    \( V_1 + V_2 = V_0 \) $\rightarrow$ \( V_2 \) is the shaded area.\\
    Hence\\
    \texttt{Sum} \( V_1 \) \( V_2 \) \( V_0 \) $\rightarrow$ \( V_1 + V_2 = V_0 \) $\rightarrow$ \( V_2 = V_0 - V_1 \)\\
    5. Extract the shaded‐region area\\
    Get \( V_2 \)\\\\
    \boxed{\texttt{Circle\_R\_Area N0 V0 RNgon\_L\_Area C6 N0 V1 Sum V1 V2 V0 Get V2}} \\
    \boxed{ N0 = 7 }
    \tcbline

    \textbf{Answer:}\\
    The solver executes the program and outputs 26.632.  \textcolor{green}{\faCheck}  

\end{tcolorbox}


\begin{center}
\begin{minipage}{1.0\linewidth} 
    \begin{tcolorbox}[colback=white, colframe=DarkBlue, title=\textbf{Reasoning Error Case},arc=3mm,boxrule=0.8pt]
    \begin{minipage}{0.7\linewidth}
        \textbf{Question:}\\[1pt]
        In \(\odot B\), \(m \angle L B M = 3x\) and \(m \angle L B Q = 4x+61\). \(M Q \perp K B\) at \(B\), what is the measure of $\angle P B Q$ ?
    \end{minipage}%
    \hspace{10mm} 
    \begin{minipage}{0.2\linewidth}
        \centering
        \includegraphics[width=\linewidth,height=2cm]{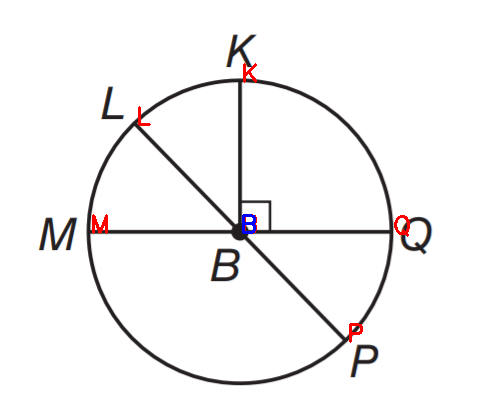} 
    \end{minipage}
\end{tcolorbox}

\begin{tcolorbox}[colback=white, title=\textbf{Natural Language Solution}, arc=3mm, boxrule=0.8pt, left=4mm, right=4mm, top=1.5mm, bottom=1.5mm, height=10cm]
    \textbf{Model's Response:}

    Since \(MQ \perp KB\) at \(B\), \textcolor{red}{ \(KB\) is the perpendicular bisector of \(LM\) at \(B\)}....\(m \angle LBM = 3x\) and \(m \angle LBQ = 4x+61\)...
    Hence, we can write the equation:
    $
    m \angle L B M = m \angle Q B M
    $
    
    Using the equation \(3x = 4x + 61 - 3x\):
    
    $
    x = -61
    $
    
    Since \(x\) cannot be negative, let us use the properties of circle :
    
    m $\angle L B Q = 180-3x$.
    
    Using $\angle L B Q = 4x+61 = 4x+61$.
    
    $\angle P B Q = 90-(4x+61)$.
    
    Using :
    
    m $\angle L B Q = 4x+61 = 4x+61$.
    
    ...
    
    \textcolor{red}{$\angle PBQ = 4x+61 = 4 * 13+61 = 4 * 13 = 52$.}
    
    $\angle PBQ = \angle Q B M = 4x+61 =52$.

    $\angle PBQ = \angle QBM = \angle PBQ = \angle PBQ = \angle PBQ = \angle PBQ = ... $

    \tcbline
    \vspace{5pt}
    \textbf{Answer:}\\
    52 (Correct: 51) \textcolor{red}{$\boldsymbol{\times}$} 
    
    \vspace{5pt}
    \textbf{Error:} \\
    The reasoning process involves random guessing, incorrectly assumes the value of x to be 13, and ultimately falls into an infinite loop.
    \end{tcolorbox}

\begin{tcolorbox}[colback=white, colframe=DarkGreen, title=\textbf{Formal-Integrated Solution}, arc=3mm,
boxrule=0.8pt,left=4mm, right=4mm, top=1.5mm, bottom=1.5mm,height=10cm]
    \textbf{Model's Response:}

    Step-by-step reasoning: \\
    1. Let \\
    \( N_0 = m\angle LB M = 3x \) \\
    \( N_1 = m\angle LBQ = 4x+61 \) \\
    2. Since $\angle$LB M and $\angle$LBQ are supplementary on a straight line, \\
    \( N_0 + N_1 = 180 \). \\
    Use \texttt{Sum} \( N_0 \) \( N_1 \) C180 $\rightarrow$ \( N_0 + N_1 = 180 \) (3x + (4x + 61) = 180) \\
    3. $\angle$PBQ and $\angle$LBQ are supplementary on the straight line through the diameter, so \\
    \( m\angle PBQ = 180 - m\angle LBQ \). \\
    Hence \\
    \( V_0 = 180 - N_1 \). \\
    Use \texttt{Sum} \( N_1 \) \( V_0 \) C180 $\rightarrow$ \( N_1 + V_0 = 180 \) $\rightarrow$ \( V_0 \) is \( m\angle PBQ \) \\
    4. Extract the value of $\angle$PBQ \\
    \texttt{Get} \( V_0 \) \\\\
    \boxed{\texttt{Sum N0 N1 C180 Sum N1 V0 C180 Get V0}} \\
    \boxed{\texttt{N0 = 3*x  N1 = 4*x + 61}}
    \vspace{1pt}
    \tcbline
    \textbf{Answer:}\\[1pt]
    The solver executes the program and outputs 51.  \textcolor{green}{\faCheck}  
\end{tcolorbox}
    
\end{minipage}
\end{center}

\subsection{Training on Larger Base Model}
\label{appendix:larger_model}

To investigate whether this framework is scalable to larger models, we conduct additional experiments using Qwen-2.5-VL-32B as our base model. The results are shown in~\cref{tab:large_model}, it can be observed that using a larger base model, the Pass@8 performance increases from 77.1 to 84.6. This demonstrates that our training framework effectively scales with model capacity.

\begin{table}[t]
\centering
\caption{SFT initialization results on a larger base model.}
\label{tab:large_model}
\begin{tabular}{lcccc}
\toprule
\textbf{Base Model} & \textbf{Pass@1} & \textbf{Pass@2} & \textbf{Pass@4} & \textbf{Pass@8}\\
\midrule
Qwen-2.5-VL-7B & 42.2 &  55.3 & 67.2 & 77.1 \\
Qwen-2.5-VL-32B & 56.5 & 67.9 & 77.3 & 84.6 \\
\bottomrule
\end{tabular}
\end{table}

\begin{table}[t]
\centering
\caption{Compare with o4-mini varying number of maximum tokens.}
\label{tab:compare_o4}
\begin{tabular}{lcccc}
\toprule
\textbf{Methods} & \textbf{Max Tokens} & \textbf{Accuracy} \\
\midrule
\multirow{3}{*}{o4-mini} & 2048 & 53.3\\
& 4096 & 70.4\\
& 8192 & 80.2\\
\midrule
GF-Reasoner (Ours) & 1024 & 68.7 \\
\bottomrule
\end{tabular}
\end{table}

\begin{table}[t!]
\caption{Comparison with existing methods in geometry problem solving.}
\label{tab:compare_existing}
\centering
\begin{tabular}{c|ccccccc}
\toprule
Method &  Language & CoT Reasoning   \\ \hline 
Specialist Geometry System & Formal Language & $\times$ \\
MLLMs & Natural Language  & \checkmark \\ 
GF-Reasoner (Ours)  & Natural-Formal Interleved & \checkmark \\ 
\bottomrule
\end{tabular}
\end{table}

\subsection{Comparison with OpenAI o4-mini}
\label{appendix:compare_o4}

We conduct a comparative analysis with the state-of-the-art multimodal model, OpenAI's o4-mini. As shown in~\cref{tab:compare_o4}, our model achieves comparable accuracy with o4-mini when the maximum token limit is 4096. Notably, o4-mini exhibits improved performance on the PGPS9k-test benchmark as its maximum token count increases, demonstrating the effectiveness of test-time scaling. These results motivate future exploration of long-chain reasoning (long-CoT) in the formal-integrated framework.

\subsection{Paradigm Comparison with Existing Solutions}
\label{appendix:compare_paradigm}

We provide a paradigm comparison between our approach and existing solutions in~\cref{tab:compare_existing}. Specialist geometry systems(e.g., GeoX, PGPSNet) solely rely on formal languages and lack CoT reasoning capabilities. Multimodal large language models(MLLMs, e.g., QwenVL, Vision-R1) exhibit reasoning capacity but are constrained to pure natural language. Our GF-Reasoner uniquely bridges this gap through a natural-formal interleaved reasoning paradigm, combining the precision of formal systems with the flexibility of natural language reasoning.

%% file: appendix/experiment_details.tex
\section{Additional Experiment Details}

\subsection{Supervised Fine-tuning}

\textbf{Datasets.} Supervised Fine-Tuning is performed on the formal-integrated Chain-of-Thought (CoT) dataset synthesized from three geometry problem-solving benchmarks: PGPS9k, UniGeo, and Geo170k. The final dataset comprises 11,000 high-quality synthetic reasoning chains, with contributions of 5,000 from PGPS9k, 2,200 from UniGeo, and 3,800 from Geo170k.

\textbf{Training Details.} We fine-tune the Qwen-2.5-VL-7B-Instruct model implemented via LLaMA-Factory framework\footnote{https://github.com/hiyouga/LLaMA-Factory}. The key training parameters are as follows:

\begin{verbatim}
torchrun --nproc_per_node 4 src/train.py \
    --finetuning_type full \
    --do_train \
    --adam_beta2 0.95 \
    --model_name_or_path Qwen2.5-VL-7B-Instruct \
    --trust_remote_code \
    --preprocessing_num_workers 8 \
    --template qwen2_vl \
    --warmup_ratio 0.03 \
    --weight_decay 0.0 \
    --per_device_train_batch_size 8 \
    --gradient_accumulation_steps 4 \
    --gradient_checkpointing \
    --ddp_timeout 9000 \
    --learning_rate 2e-5 \
    --lr_scheduler_type cosine \
    --logging_steps 2 \
    --cutoff_len 4096 \
    --num_train_epochs 2 \
    --bf16 \
    --seed 42 \
    --flash_attn fa2 \
\end{verbatim}

\textbf{Compute Resources.} The SFT stage runs using 4 NVIDIA A800-80GB GPUs. The training process completes in approximately 0.5 hours.

\subsection{Reinforcement Learning}

\textbf{Datasets.} Reinforcement Learning is performed on 19.5k samples drawn from the training splits of three geometry problem-solving benchmarks: PGPS9k, UniGeo, and Geo170k. The distribution consists of 8k samples from PGPS9k, 3.5k from UniGeo, and 8k from Geo170k.

\textbf{Training Details.} This stage builds upon the model initialized from supervised fine-tuning. We employ the GRPO algorithm~\citep{shao2024deepseekmath} implemented in the verl framework\footnote{https://github.com/volcengine/verl}, with key hyperparameters configured as follows:

\begin{verbatim}

python3 -m verl.trainer.main_ppo \
    algorithm.adv_estimator=grpo \
    data.train_batch_size=256 \
    data.max_prompt_length=2048 \
    data.max_response_length=1024 \
    data.val_batch_size=5120 \
    data.filter_overlong_prompts=True \
    data.truncation=`error' \
    data.image_key=images \
    actor_rollout_ref.model.path=$MODEL_PATH \
    actor_rollout_ref.actor.optim.lr=1e-6 \
    actor_rollout_ref.actor.optim.lr_warmup_steps=0 \
    actor_rollout_ref.actor.optim.weight_decay=0.0 \
    actor_rollout_ref.model.use_remove_padding=True \
    actor_rollout_ref.actor.ppo_mini_batch_size=64 \
    actor_rollout_ref.actor.ppo_micro_batch_size_per_gpu=16 \
    actor_rollout_ref.actor.use_kl_loss=True \
    actor_rollout_ref.actor.kl_loss_coef=0.0 \
    actor_rollout_ref.actor.kl_loss_type=low_var_kl \
    actor_rollout_ref.actor.entropy_coeff=0 \
    actor_rollout_ref.actor.clip_ratio_high=0.28 \
    actor_rollout_ref.model.enable_gradient_checkpointing=True \
    actor_rollout_ref.actor.fsdp_config.param_offload=False \
    actor_rollout_ref.actor.fsdp_config.optimizer_offload=False \
    actor_rollout_ref.rollout.log_prob_micro_batch_size_per_gpu=64 \
    actor_rollout_ref.rollout.tensor_model_parallel_size=1 \
    actor_rollout_ref.rollout.name=vllm \
    actor_rollout_ref.rollout.gpu_memory_utilization=0.8 \
    actor_rollout_ref.rollout.enable_chunked_prefill=False \
    actor_rollout_ref.rollout.enforce_eager=False \
    actor_rollout_ref.rollout.free_cache_engine=False \
    +actor_rollout_ref.rollout.enable_prefix_caching=True \
    actor_rollout_ref.rollout.n=$n_samples_per_prompt \
    actor_rollout_ref.rollout.temperature=1.0 \
    actor_rollout_ref.rollout.val_kwargs.temperature=0.0 \
    actor_rollout_ref.rollout.val_kwargs.n=1 \
    actor_rollout_ref.rollout.val_kwargs.do_sample=False \
    actor_rollout_ref.ref.log_prob_micro_batch_size_per_gpu=64 \
    actor_rollout_ref.ref.fsdp_config.param_offload=True \
    algorithm.kl_ctrl.kl_coef=0.0 \
    custom_reward_function.path=./verl/utils/reward_score/pgps9k.py \
    custom_reward_function.name=compute_score \
    trainer.val_before_train=True \
    trainer.critic_warmup=0 \
    trainer.n_gpus_per_node=8 \
    trainer.nnodes=1 \
    trainer.total_epochs=15 \
    
\end{verbatim}

\textbf{Compute Resources.} The RL training runs using 8 NVIDIA H20-96GB GPUs. The training process completes in approximately 83 hours.